\documentclass{article}

\PassOptionsToPackage{numbers, compress}{natbib}
\usepackage[preprint]{neurips_2023}

\usepackage[utf8]{inputenc} 
\usepackage[T1]{fontenc}    
\usepackage{url}            
\usepackage{booktabs}       
\usepackage{amsfonts}       
\usepackage{nicefrac}       
\usepackage{microtype}      
\usepackage[table,xcdraw]{xcolor}         
\usepackage{graphicx}
\usepackage{subcaption}
\usepackage{algorithmic}
\usepackage{algorithm}
\usepackage{multirow}
\usepackage{bbding}
\usepackage{setspace}
\usepackage{wrapfig}
\usepackage{fancybox}
\usepackage[pagebackref=true,breaklinks=true,colorlinks,urlcolor=black,bookmarks=false]{hyperref}

\def\datasetname{AS-1B}

\definecolor{mygray}{gray}{.92}

\title{The All-Seeing Project: Towards Panoptic Visual Recognition and Understanding of the Open World}

\author{
 \textbf{Weiyun Wang$^{*1,2}$, Min Shi$^{*1,3}$, Qingyun Li$^{*1,4}$, Wenhai Wang$^{*1}$, Zhenhang Huang$^{*1}$}, \\ 
 \textbf{Linjie Xing$^{*1}$, Zhe Chen$^{1,5}$, Hao Li$^{1,6}$, Xizhou Zhu$^{1,7}$, Zhiguo Cao$^3$, Yushi Chen$^4$}, \\
 \textbf{Tong Lu$^{5}$, Jifeng Dai$^{\dagger 1,8}$, Yu Qiao$^{1}$} \\ 
	$^1$OpenGVLab, Shanghai AI Laboratory \quad $^2$Fudan University \\
    $^3$Huazhong University of Science and Technology \quad $^4$Harbin Institute of Technology \\ $^5$Nanjing University \quad
	$^6$The Chinese University of Hong Kong \\ $^7$SenseTime Research \quad $^8$Tsinghua University \\\\
{Code: \url{https://github.com/OpenGVLab/all-seeing}}\\
{Demo: \url{https://huggingface.co/spaces/OpenGVLab/all-seeing}}
}

\definecolor{gray9}{gray}{.9}
\definecolor{gray95}{gray}{.95}

\definecolor{gray8}{gray}{.8}
\definecolor{gray85}{gray}{.85}
\def\ie{\emph{i.e.}}
\def\eg{\emph{e.g.}}

\newcommand\blfootnote[1]{%
\begingroup
\renewcommand\thefootnote{}\footnote{#1}%
\addtocounter{footnote}{-1}%
\endgroup
}
\begin{document}

\maketitle
\thispagestyle{empty}
\blfootnote{\noindent $^{*}$Equal contribution. This work is done when Weiyun Wang, Min Shi, and Qingyun Li are interns at Shanghai AI Laboratory.
$^{\dagger}$Corresponding to Jifeng Dai <daijifeng@tsinghua.edu.cn>.
}

\begin{abstract}

We present the All-Seeing (AS)\footnote{``All-Seeing'' is derived from ``The All-Seeing Eye'', which means having complete knowledge, awareness, or insight into all aspects of existence.} project: a large-scale data and model for recognizing and understanding everything in the open world.
Using a scalable data engine that incorporates human feedback and efficient models in the loop,
we create a new dataset (\datasetname) with over 1 billion regions annotated with semantic tags, question-answering pairs, and detailed captions.
It covers a wide range of 3.5 million common and rare concepts in the real world, and has 132.2 billion tokens that describe the concepts and their attributes.
Leveraging this new dataset, we develop the All-Seeing model (ASM), a unified framework for panoptic visual recognition and understanding. The model is trained with open-ended language prompts and locations, 
which allows it to generalize to various vision and language tasks with remarkable zero-shot performance, including region-text retrieval, region recognition, captioning, and question-answering. 
We hope that this project can serve as a foundation for vision-language artificial general intelligence research. 
\end{abstract}

\section{Introduction}

\begin{figure}[t]
\hsize=\textwidth
\centering
\begin{subfigure}{0.31\textwidth}
    \centering
    \includegraphics[width=1.0\textwidth]{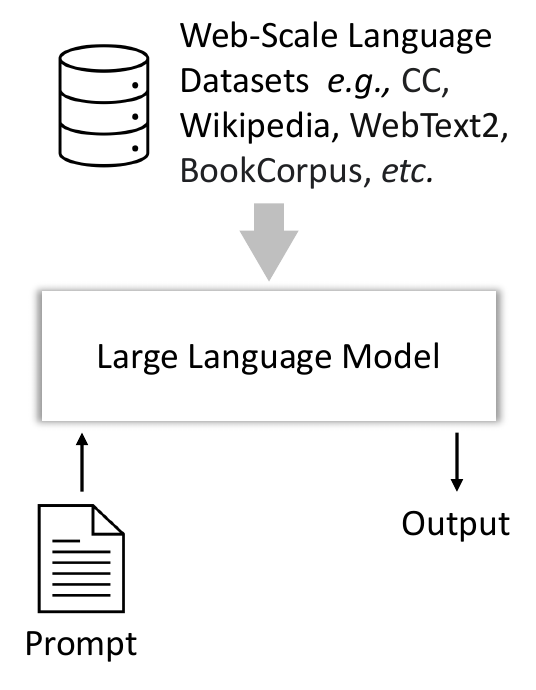}
    \caption{
    Large Language Models (LLMs) possess extensive world knowledge and demonstrate impressive reasoning capabilities, but lack the ability to receive and comprehend visual information.
    }
    \label{fig:1a}
\end{subfigure}    
\hspace{0.1in}
\begin{subfigure}{0.31\textwidth}
     \centering
     \includegraphics[width=1.0\textwidth]{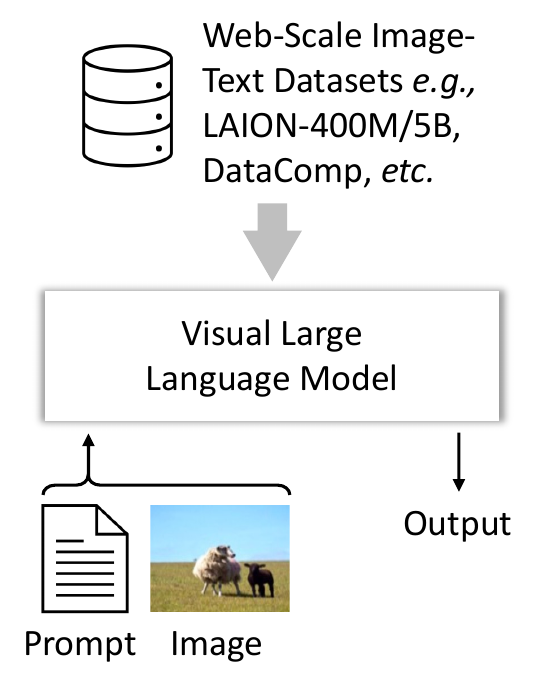}
     \caption{
    Visual Large Language Models (VLLMs) can process both text and images, but they can only capture the holistic visual information of the whole image and understand it based on LLMs.
     }
     \label{fig:1b}
\end{subfigure}
\hspace{0.1in}
\begin{subfigure}{0.31\textwidth}
     \centering
     \includegraphics[width=1.0\textwidth]{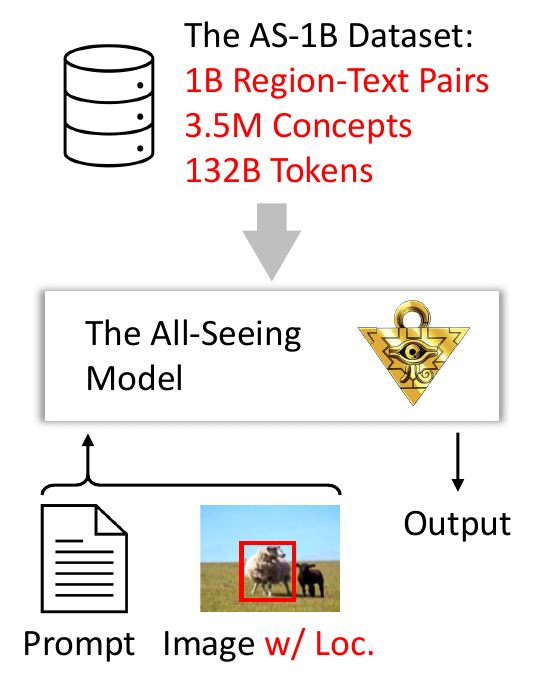}
     \caption{
     Our All-Seeing Model (ASM) can comprehensively recognize and understand the objects or concepts in regions of interest, while maintaining the capabilities of VLLMs and LLMs.
     }
     \label{fig:1c}
\end{subfigure}
\caption{
\textbf{Overview and comparison of our All-Seeing project with other popular large foundation models.} To address the limitations of LLMs in understanding visual inputs and VLLMs in effectively leveraging region-aware information, we propose (1) a large-scale dataset \datasetname\, which consists of 2 billion region-text pairs, 3.5 million open-world concepts, and over 1 billion tokens of region-related question-answering and caption; and
(2) the All-Seeing model (ASM), which is capable of recognizing and understanding context in arbitrary regions.
}
\label{fig:cmp}
\end{figure}

Creating artificial general intelligence (AGI) systems that can match human intelligence and excel in any task across domains is the ultimate goal of artificial intelligence.
Recent advancements in Large Language Models (LLMs) have demonstrated impressive zero-shot capabilities in user-tailored natural language processing (NLP) tasks, suggesting new avenues for achieving AGI.
However, as shown in Fig.~\ref{fig:1a}, most popular LLMs~\cite{openai2022chatgpt,ouyang2022instruct-tuning, taori2023alpaca, 2023internlm, touvron2023llama, moss, chiang2023vicuna} are limited to processing language information and lack the ability to perceive or understand the visual world.

Although there have been some recent developments~\cite{openai2023gpt4,zhu2023minigpt-4,liu2023llava,li2022blip,li2023blip-2,instructblip,ye2023mplug-owl,liu2023interngpt} in open-world visual understanding, they are primarily focused on understanding images as a whole, rather than recognizing and comprehending individual instances within the scene (see Fig.~\ref{fig:1b}).
\emph{This goes against 
the nature of the human visual system}, as described by the feature integration theory~\cite{TREISMAN198097}, which suggests that we attentively gather visual features and contexts in certain regions to achieve high-level understanding and recognition, rather than analyzing all information simultaneously.

To achieve instance-level visual understanding like humans, there are two major challenges as follows:
(1) \emph{The scarcity of open-world instance-text pair data.} As listed in Table~\ref{tab:dataset-comparisons}, existing datasets, such as Visual Genome~\cite{krishna2017visualgenome}, have limitations in terms of data scale and open-world annotations. Laion-5B~\cite{schuhmann2022laion5b} only contains web-crawled image-text pairs without location information, and SA-1B~\cite{kirillov2023segment} lacks semantic information.
(2) \emph{The lack of spatial information modeling in most existing models}. These models mainly focus on whole-image understanding as mentioned above.

In this work, we propose the All-Seeing (AS) project for open-world panoptic visual recognition and understanding, driven by the goal of creating a vision system that mimics human cognition. The term ``panoptic'' refers to including everything visible in one view~\cite{kirillov2019panoptic}. The AS project addresses the challenges from both the data and model perspectives.

\begin{figure}[!t]
	\centering
        \includegraphics[width=1.0\linewidth]{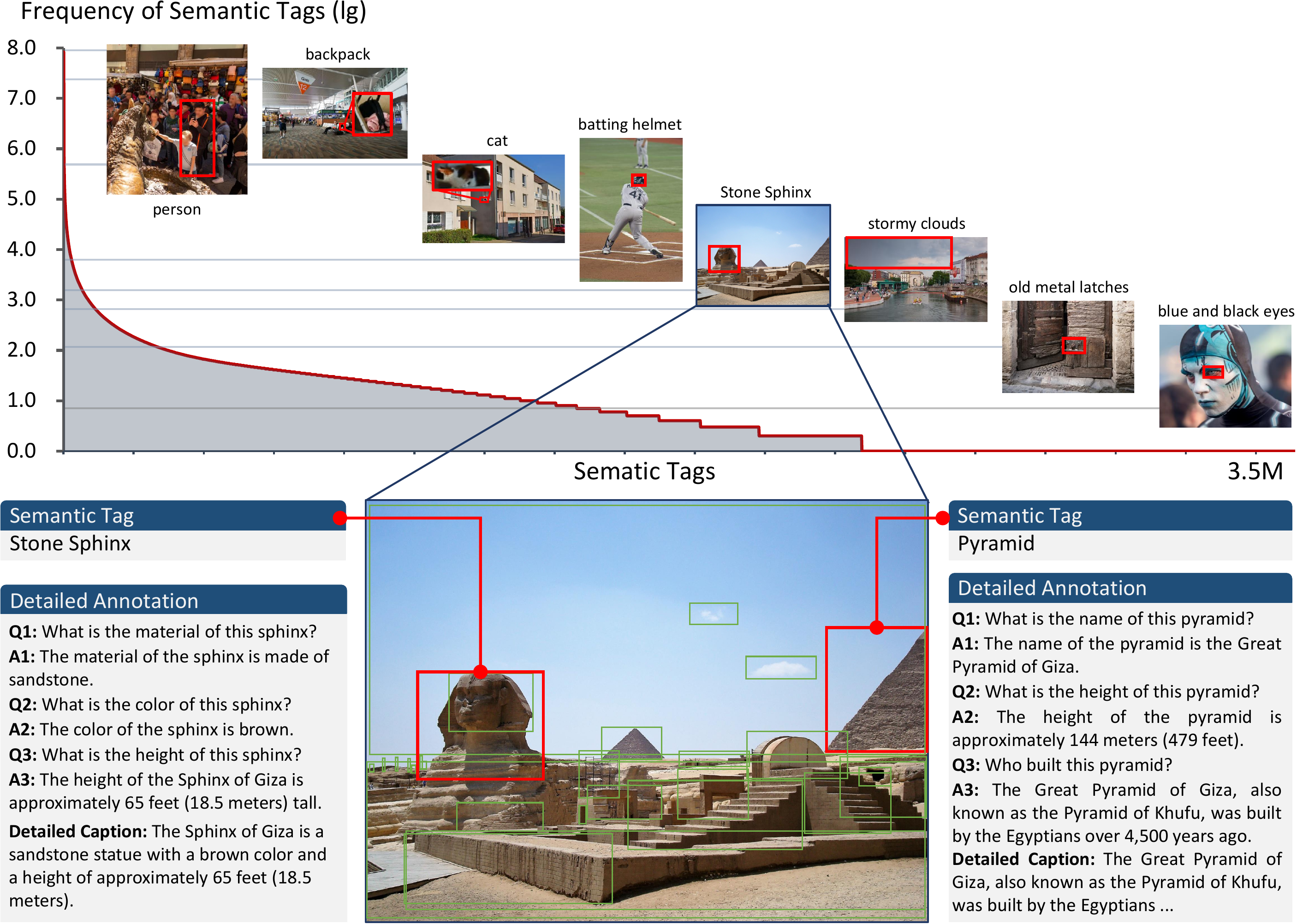}
    \caption{\textbf{Semantic concepts and annotations in the \datasetname\ dataset}. The semantic tags in \datasetname\ dataset encompass a wide range of concepts, from common objects to rare and fine-grained categories with attributes. Beyond brief semantic tags, detailed annotations, including visual-question-answering pairs and region captions are also provided.}
    \label{fig:semantic-tag-distribution-and-display}
\end{figure}

\textbf{From the data aspect}, we propose the All-Seeing 1B (\datasetname) dataset, consisting of over 1 billion region annotations in various formats, such as semantic tags, locations, question-answering pairs, and captions (refer to Fig.~\ref{fig:semantic-tag-distribution-and-display}). 
\datasetname\ dataset is made possible by a scalable semi-automatic data engine, which significantly lowers the previously unaffordable expense of manually annotating a massive amount of open-world semantics.
The data engine operates in a ``data-human-model'' loop, iteratively refining data quality. 
Initially, diverse models, including large language models (LLMs) \cite{chiang2023vicuna}, detection\cite{wang2022internimage,fang2023eva02,li2022glip}, captioning~\cite{li2022blip}, and visual question answering models (VQA) \cite{liu2023llava,zhu2023minigpt-4,liu2023interngpt}, are employed as ``annotators'', which add semantic annotations to dense region proposals generated by state-of-the-art object detectors~\cite{kirillov2023segment,fang2023eva02,li2022glip,wang2022internimage}. 
Subsequently, human annotators verify the generated pseudo labels and provide feedback with high-quality data, which is then used to fine-tune the models to improve their performance. The enhanced models are then used to re-annotate the data, starting another iteration of the loop.
As shown in Fig.~\ref{fig:semantic-tag-distribution-and-display}, \datasetname\ contains a wide range of open-world concepts, including over 3.5 million different semantic tags ranging from common categories (\eg, human, backpack) to fine-grained or rare categories with attributes (\eg, old metal latches).
\datasetname\ also encompasses the 3.3 billion visual question-answering pairs and 1.2 billion region captions for 1.2 billion regions.

\textbf{In terms of the model perspective}, we propose the All-Seeing model (ASM), a unified location-aware image-text foundation model. The model consists of two key components: a location-aware image tokenizer and an LLM-based decoder. The location-aware image tokenizer uses location information such as box, mask, and point set as conditions (see Fig.~\ref{fig:1c}) to extract image features, which contribute to the location capability of ASM. 
The LLM-based decoder inherits the world knowledge and reasoning capability from LLMs such as LLaMA \cite{touvron2023llama}, providing a strong foundation for visual recognition and understanding.
In addition, to unify image-text aligning and generation tasks, we introduce a new decoding approach, where the aligning tasks are reformulated as a ``special'' generation task, enabling our model to generalize to various vision-language tasks with shared weights.

Compared to previous methods~\cite{radford2021clip,alayrac2022flamingo,li2022blip,liu2023llava,zhu2023minigpt-4}, our work offers several advantages as follows: (1) Our model not only excels in image-level understanding but also demonstrates exceptional capability in recognizing and comprehending objects at the instance level, closely aligning with human cognitive processes. 
(2) Our model is a unified framework that supports a wide range of image-text tasks, including discriminative tasks like image-text retrieval, as well as generative tasks such as visual captioning and question-answering.
(3) Our model comes with \datasetname\, the largest dataset with open-world panoptic semantics. Data and models feed each other in the data engine, iteratively improving the model performance, data scale and diversity.

In summary, our contributions are three folds:

(1) We propose a new large-scale dataset (\datasetname) for open-world panoptic visual recognition and understanding, using an economical semi-automatic data engine that combines the power of off-the-shelf vision/language models and human feedback. As reported in Table~\ref{tab:dataset-comparisons}, we have 159 times more semantic tags and 33 times more regions compared with its counterparts. 

(2) Based on the dataset, we develop a unified vision-language foundation model (ASM) for open-world panoptic visual recognition and understanding. Aligning with LLMs, our ASM supports versatile image-text retrieval and generation tasks, demonstrating impressive zero-shot capability.

(3) We evaluate our model on a representative vision and vision-language tasks. 
Our ASM outperforms CLIP~\cite{radford2021clip} by 10.4 and 14.3 (mAP) on COCO~\cite{lin2014microsoft} and LVIS~\cite{gupta2019lvis} in zero-shot region recognition tasks. 
When trained with \datasetname\ (region-level data) and LaionCOCO~\cite{laioncoco} (image-level data), our model achieves superior zero-shot and fine-tuned performance compared to recent image-level~\cite{li2023blip-2,instructblip,wang2023visionllm,yu2022coca,huang2023kosmos-1} and region-level~\cite{yu2017slr,wu2022grit,peng2023kosmos2} VLLMs.

\begin{table}[!t]
        \centering
        \small
	\begin{tabular}{l|cccc|cc}
  \toprule
            Dataset & \#Images & \#Regions & \#Concepts & \#Tokens & Location & Semantic \\
            \midrule
            \textcolor{gray}{\emph{Image-Level}} & &&& \\
		ImageNet-22K~\cite{deng2009imagenet}                & 15M     & $-$        & 22,000  & $-$     & $-$            & Closed-Set                  \\
        COCO Caption~\cite{chen2015coco-caption}            & 0.1M    & $-$        & $-$     & 8.4M    & $-$            & Closed-Set                  \\
        SBU~\cite{ordonez2011sbu}                           & 0.8M    & $-$        & $-$     & 14.6M   & $-$            & Open-World                  \\
        CC12M~\cite{changpinyo2021cc12m}                    & 12.4M   & $-$        & $-$     & 250.9M  & $-$            & Open-World                  \\
        YFCC15M~\cite{yfcc15m}                              & 15M     & $-$        & $-$     & 1.0B    & $-$            & Open-World                  \\
        COYO700M~\cite{kakaobrain2022coyo700m}              & 700M    & $-$        & $-$     & 15.0B   & $-$            & Open-World                  \\
        Laion-5B~\cite{schuhmann2022laion5b}                & 5B      & $-$        & $-$     & 135.0B  & $-$            & Open-World                  \\
  \midrule
  \textcolor{gray}{\emph{Class-Agnostic}}& &&& \\
        SA-1B~\cite{kirillov2023segment}                    & 11M     & 1.1B       & $-$     & $-$     & Open-World     & $-$                        \\
  \midrule
  \textcolor{gray}{\emph{Region-Level}}& &&& \\
		COCO~\cite{lin2014microsoft}                      & 0.1M  & 0.9M   & 80     &  $-$     & Closed-Set                    & Closed-Set                    \\
		LVIS~\cite{gupta2019lvis}                         & 0.1M  & 1.5M   & 1,203  &  $-$     & Closed-Set                    & Closed-Set                    \\
		Objects365~\cite{shao2019objects365}              & 0.6M  & 10.1M  & 365    &   $-$    & Closed-Set                    & Closed-Set                    \\
		Open Images~\cite{kuznetsova2020openimages}       & 1.5M  & 14.8M  & 600    &  $-$     & Closed-Set                    & Closed-Set                    \\
		BigDetection~\cite{cai2022bigdetection}           & 3.5M  & 36.0M  & 600    &  $-$     & Closed-Set                    & Closed-Set                    \\
		V3Det~\cite{wang2023v3det}                        & 0.2M  & 1.5M   & 13,029 & $-$      & Closed-Set                    & Closed-Set                    \\
		Visual Genome~\cite{krishna2017visualgenome}      & 0.1M  & 0.3M   & 18,136 &  51.2M   & Open-World                    & Open-World                    \\
  \rowcolor{mygray}
  AS-1B (ours)                    & \textbf{11M}    & \textbf{1.2B}       & \textbf{3.5M}     & \textbf{132.2B}   & Open-World                      & Open-World                      \\ 
  \bottomrule
	\end{tabular}
        \setlength{\abovecaptionskip}{0.15cm}
        \caption{
        \textbf{Comparison with popular vision and vision-language datasets}. 
        ``\#'' denotes the number of something. We see that the proposed AS-1B dataset has a significantly larger data scale and diversity than prior region-level datasets.
        }
        \label{tab:dataset-comparisons}
\end{table}

\section{Related Work}

\textbf{The Emergence of Large Language Models.}
In recent years, based on the large-scale text corpora~\cite{pile,ccnews,radford2019gpt2,bookcorpus,Stories,wudao_corpora}, the field of Large Language Models (LLMs) has witnessed remarkable progress \cite{radford2019gpt2,brown2020gpt3, longpre2023flan,raffel2020t5,zhang2022opt}. 
Prominent models such as ChatGPT \cite{openai2022chatgpt} and GPT-4 \cite{openai2023gpt4} have demonstrated excellent performance across various tasks, showcasing their potential for semantic understanding, dialogue generation, programming, mathematical problem-solving, and more.
However, there is a growing concern that these leading institutes are becoming increasingly conservative in sharing the technical details of their models and roadmaps.
To catch up with the performance of ChatGPT, the open-source community has devoted substantial efforts~\cite{touvron2023llama, wang2022self_instruct, taori2023alpaca, chiang2023vicuna,zeng2022glm, gao2022pal,zhang2023llamaadapter}.
For instance, Self-Instruct \cite{wang2022self_instruct} introduced an iterative bootstrapping algorithm that leverages off-the-shelf LLMs and a seed set of manually-written instructions to expand the instruction collection. 
Alpaca \cite{taori2023alpaca} utilized the Self-Instruct technique to generate high-quality instruction-following data, which was then used to fine-tune the LLaMA \cite{touvron2023llama} model.
Vicuna \cite{chiang2023vicuna} demonstrated that fine-tuning on user-shared ChatGPT conversations can spark dialog and improve instruction-following capabilities.
Furthermore, there has been a focus on improving multilingual capabilities, particularly in Chinese, with LLMs like Chinese-Alpaca \cite{cui2023chinesellama}, GLM-130B \cite{zeng2022glm}, InternLM \cite{2023internlm}, MOSS \cite{moss}, and others.
These LLMs have shown excellent proficiency in learning world knowledge, which lays the groundwork for open-world understanding.

\textbf{Datasets for Visual Recognition and Understanding.}
The dataset plays a critical role in the advancement of deep learning models, especially in the field of visual recognition and comprehension.
Prior to the era of large-scale models, datasets are primarily closed-world or have limited data scale, including CIFAR-10/100~\cite{krizhevsky2009cifar}, ImageNet~\cite{deng2009imagenet}, and iNaturalist~\cite{van2018inaturalist} for image classification, Pascal VOC~\cite{everingham2010pascal}, COCO~\cite{lin2014microsoft}, LVIS~\cite{gupta2019lvis}, OpenImages~\cite{kuznetsova2020openimages}, ADE20K~\cite{zhou2019ade20k}, and Cityscape~\cite{cordts2016cityscapes} for visual location, as well as SBU~\cite{ordonez2011sbu}, CC3M~\cite{sharma2018cc3m}, CC12M~\cite{changpinyo2021cc12m}, YFCC15M~\cite{thomee2016yfcc100m}, and VQA~\cite{antol2015vqa}, VQA 2.0~\cite{goyal2017making}, ICDAR 2015~\cite{karatzas2015icdar}, SCUT-CTW1500~\cite{yuliang2017detecting}
for visual understanding.
Additionally, datasets like Visual Genome~\cite{krishna2017visualgenome} and Visual7W~\cite{visual7w}
integrate visual location and understanding, offering more comprehensive tasks to describe the visual world. 
However, these datasets have limited semantics and fail to encompass diverse scenarios in the open world, which hinders the generalization ability of models.
To achieve open-world capability, CLIP~\cite{radford2021clip} and ALIGN~\cite{jia2021align} propose training models using web-scale image-text pairs collected from the internet. Subsequent works, such as Laion-400M~\cite{schuhmann2021laion400m}, Laion-5B~\cite{schuhmann2022laion5b}, COYO-700M~\cite{kakaobrain2022coyo700m} and DataComp~\cite{gadre2023datacomp}, have also been introduced for open-source research.
However, these approaches only include descriptions or question-answering pairs corresponding to the entire image, resulting in models struggling to accurately recognize and understand specific objects at the instance level.
Recently, Kirillov et al. introduced SA-1B \cite{kirillov2023segment}, which provides open-world location information such as boxes and masks but still lacks semantic details. 
So existing datasets cannot meet the requirements of data scale, open-world location and semantics necessary for achieving visual AGI models, thus posing challenges in supporting human-like panoptic visual recognition and understanding.

\textbf{Models for Visual Recognition and Understanding.}
Significant advancements have been made in the field of visual recognition and understanding in recent years.
Previous methods~\cite{he2017maskrcnn,kamath2021mdetr,chen2022vision,zhu2021deformable,chen2022diffusiondet,kirillov2019panoptic,xie2020polarmask,li2023mask} mainly concentrate on the close-set recognition while recent works begin to focus on the open world understanding.
Models trained with contrastive learning-based methods, including CLIP~\cite{radford2021clip}, ALIGN~\cite{jia2021align}, EVA~\cite{fang2023eva} and FLIP~\cite{li2023flip}, are able to recognize and understand the open world semantics under an image-text matching framework while the lack of generation ability limits their applicability.
To address this limitation, subsequent works, such as SimVLM~\cite{wang2021simvlm}, UniPerceiver~\cite{zhu2022uni_p}, VL-BERT~\cite{bao2022vlbert}, VLMo~\cite{bao2022vlmo}, BEiT-3~\cite{wang2022beit3}, ALBEF~\cite{li2021albef}, CoCa~\cite{yu2022coca}, as well as Flamingo~\cite{alayrac2022flamingo}, have incorporated generative training tasks. 
However, these models are trained from scratch and do not capitalize on the powerful perception capabilities of existing powerful vision foundation models for image, and Large Language Models for text, increasing the cost of developing new models.
The recent progress of LLMs~\cite{openai2023gpt4,openai2022chatgpt,radford2018gpt1,radford2019gpt2,brown2020gpt3} initiates a new era, leading to the emergency of many LLM-based multimodal models~\cite{li2022blip, li2023blip-2, zhu2023minigpt-4, liu2023llava, ye2023mplug-owl, zhang2023llamaadapter, embodiedgpt, wang2023visionllm, chen2023videollm} and interactive systems~\cite{Yang2023MMREACTPC, liu2023interngpt, shen2023hugginggpt, zhu2023chatcaptioner, li2023videochat, GITM, yao2022react}.
However, these works are only capable of recognizing the entire image, lacking the ability to comprehend specific regions within the image.
Some concurrent methods, such as ChatSpot~\cite{zhao2023chatspot}, Shikra~\cite{chen2023shikra}, KOSMOS-2~\cite{peng2023kosmos2}, and GPT4RoI~\cite{zhang2023gpt4roi} begin to focus on location-aware understanding. 
However, without the support of large-scale instance-level visual understanding data, the generalization ability of these models is still limited. Besides, these models only support generative tasks, limiting their application to discriminative tasks, such as image-text retrieval and zero-shot object recognition.
In this work, we propose a unified location-aware image-text foundation model, based on ViT-g~\cite{fang2023eva}
and Husky~\cite{liu2023interngpt}. Our model supports both image-text matching and generation tasks, expanding its range of applications and contributing to the advancement of AGI models.

\section{The All-Seeing Dataset (\datasetname)} \label{sec:dataset}
In this section, we introduce the All-Seeing-1B (\datasetname) dataset for open-world panoptic visual recognition and understanding. The dataset consists of 1.2 billion regions in 11 million images\footnote{Images source from SA-1B~\cite{kirillov2023segment}}. Each region is annotated with comprehensive information, including categories, locations, attributes, captions, and question-answer pairs. Compared with the previous visual recognition datasets like ImageNet~\cite{deng2009imagenet} and COCO~\cite{lin2014microsoft}, visual understanding datasets like Visual Genome~\cite{krishna2017visualgenome} and Laion-5B~\cite{schuhmann2022laion5b}, \emph{the proposed \datasetname\ dataset stands out due to its rich and diverse instance-level location annotation and corresponding detailed object concepts and descriptions.}

\subsection{Data Annotation Engine}

\begin{figure}[!t]
	\centering
        \includegraphics[width=1.0\linewidth]{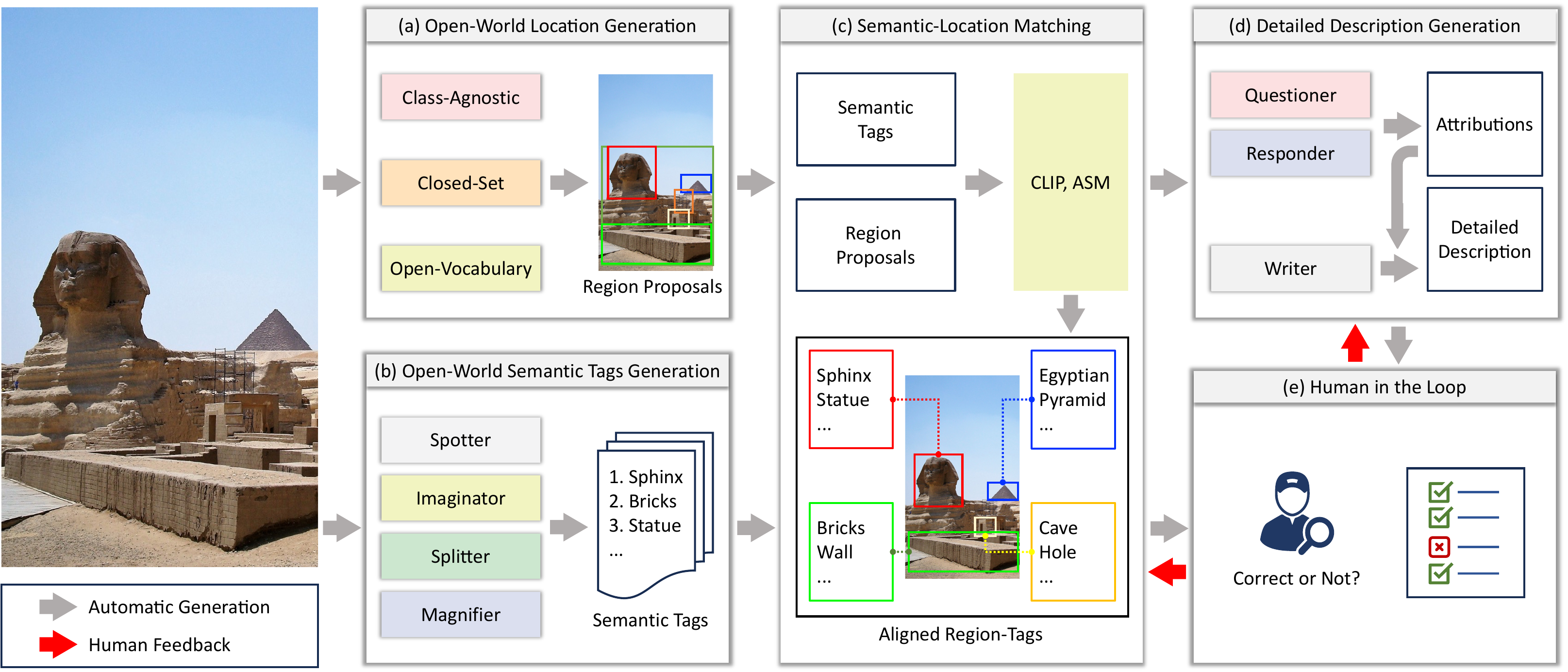}
    \caption{\textbf{Data engine for \datasetname\ dataset}. Our data engine consists of an automatic annotation pipeline (\ie, (a), (b), (c), (d)) and human verification stage (\ie, (e)). We combine strong object detectors, LLMs, and VLLMs to produce open-world locations and annotations for different regions. The automatic annotations are sampled and verified by human experts. Automated annotations are used together with human validation results to train region-aware alignment and generation models, which are then used in the automated annotation pipeline to improve data quality.
    }
    \label{fig:data-engine}
\end{figure}

We develop a semi-automatic data engine that efficiently uses a wide range of state-of-the-art foundation models as annotators, reducing the enormous labeling cost to an acceptable level.
As depicted in Fig.~\ref{fig:data-engine}, the process of the data engine begins by generating noisy pseudo data using well-trained off-the-shelf foundation models from diverse fields. Subsequently, these pseudo data are iteratively refined through multiple loops with the aid of models fine-tuned on human feedback data. 
By employing this ``data-human-model'' cycle, we can generate a large number of region-level annotations with exceptional quality.

As the core component of the data engine, the pseudo data generation pipeline consists of five steps as follows: 
(1) Creating open-world location (\eg, bounding box, mask, point set) with an ensemble of state-of-the-art class-agnostic, visual grounding, and closed-set perception models~\cite{kirillov2023segment,li2022glip,wang2022internimage,fang2023eva02};
(2) Generating open-world semantic tags using the combination of image captioning models~\cite{li2022blip,zhu2023minigpt-4} and LLMs~\cite{chiang2023vicuna};
(3) Matching the semantic tags to proper regions with image-text aligning models such as CLIP~\cite{radford2021clip};
(4) Using LLM~\cite{chiang2023vicuna} and VQA models~\cite{liu2023interngpt} to generate the attributions of each region based on the matched semantic tags;
(5) Generating detailed captions based on the semantics and attributions of each region. 

\subsection{Open-World Localization} 
\label{sec:open-world-localization}
To obtain comprehensive locations of all instances in an image, we combine the results of state-of-the-art perception models from different fields, including 
(1) \textbf{class-agnostic model}: we adopt the SAM~\cite{kirillov2023segment} to provide initial proposals of most objects in view.
(2) \textbf{closed-set detection model}: we use InternImage-H~\cite{wang2022internimage} and EVA-02~\cite{fang2023eva02} trained on BigDetection~\cite{cai2022bigdetection} and LVIS~\cite{gupta2019lvis}, respectively, to detect the common-seen objects.
(3) \textbf{grounding model}: we use GLIP~\cite{li2022glip} to ground open-world semantics generated by LLMs~\cite{zhu2023minigpt-4} (see Sec.~\ref{sec:open-world-semantics}). 
All the bounding boxes are gathered together to ensure that all possible objects in view are covered.

Due to the incomparable score ranges of different models, directly using non-maximum suppression (NMS) to eliminate duplicated proposals from multiple resources is infeasible.
Therefore, we develop an effective strategy that keeps all the semantics while removing highly-overlapped regions.
As shown in Alg.~\ref{alg:region-proposal-merging}, the merging strategy works as follows:
(1) We start by initializing the result region proposal set $\mathcal{R}$ with the class-agnostic bounding boxes generated by SAM. (2) When a set of region proposals $\mathcal{R}'$ from a new source (\eg, closed-set/grounding detector) comes in, we calculate the Intersection over Union (IoU) between the regions in $\mathcal{R}'$ and $\mathcal{R}$.
(3) If the IoU between a new region $r' \in \mathcal{R}'$ and an existing region $r \in \mathcal{R}$ is greater than a threshold $T_{\rm IoU}$, the region $r'$ is removed, and its closed-set/grounding tags are appended to the tag list of the matched region $r$.
(3) Finally, the remaining low-IoU regions in $\mathcal{R}'$ along with their tags are added to $\mathcal{R}$.
By employing this strategy, we sequentially combine the results of SAM, InternImage, EVA-02 and GLIP to obtain comprehensive location information for an image.

\begin{algorithm}[t!]
	\setstretch{1.2}  
	\renewcommand{\algorithmicrequire}{\textbf{Input:}}
	\renewcommand{\algorithmicensure}{\textbf{Output:}}
	\caption{Region Proposal Merging }
	\label{alg:region-proposal-merging}
	\begin{algorithmic}[1]
    \REQUIRE \quad  \\ 
    Existing region proposals $\mathcal{R}$  \\
        New region proposals $\mathcal{R}'$  \\
        IoU threshold $T_{\rm IoU}$ \\
    \ENSURE \quad  \\
    Merged region proposals $\mathcal{R}$  \\
        
        \FOR{region $r' \in \mathcal{R}'$}
        \STATE Calculate IoU between $r'$ and proposals in $\mathcal{R}$
            \IF{maximum IoU > $T_{\rm IoU}$}
                \STATE Merge semantic tags from $r'$ into the semantic tag of corresponding regions in $\mathcal{R}$
                \STATE Delete $r'$
            \ELSE
                \STATE Add $r'$ into $\mathcal{R}$
            \ENDIF
        \ENDFOR
\end{algorithmic}  
\end{algorithm}

\subsection{Open-World Semantic} 
\label{sec:open-world-semantics}

Manually labeling billions of regions for an open-world semantic description is impractical due to the enormous cost and time required.
On the other hand, generating annotations with off-the-shelf multi-modal models is also non-trivial, as it demands sufficient world knowledge and context-related reasoning capabilities to accurately label diverse objects in the wild. 
To remedy these challenges, we draw inspiration from the recent advancements in Large Language Models (LLMs)~\cite{brown2020gpt3, touvron2023llama, moss, taori2023alpaca, chiang2023vicuna, 2023internlm, zeng2022glm} and Visual Large Language Models (VLLMs)~\cite{openai2023gpt4, liu2023llava, li2022blip, liu2023interngpt, wang2023visionllm,zhu2023minigpt-4,ye2023mplug-owl}, we leverage a series of LLMs and VLLMs as ``semantic generators'' and tap into their vast world knowledge and reasoning abilities for open-world semantic generation.
These ``semantic generators'' can be specialized for producing short semantic tags (such as category names and brief attributes) or detailed annotations (including attributes, question-answering pairs, and captions) based on specially-designed prompts.

\subsubsection{Semantic Tags}
To generate as many semantic tags as possible for a view, different instructions are employed to harness the diverse capabilities of LLMs and VLLMs, turning them into annotators with different focuses and skills. Specifically, we have (1) a \textbf{spotter}, which identifies major instances and provides an overview of the scenes, (2) a \textbf{imaginator} that leverages world knowledge to imagine plausible objects, (3) a \textbf{splitter} that divides complicated objects into parts,  as well as (4) which zooms on each region to produce region-specific candidates. These models complement each other to create a powerful system that can generate comprehensive open-world semantic tags for each region and the entire image. Here are the details of each model:

\textbf{Spotter}. 
This module aims to list the prominent and major objects present in the given image. To achieve this, we use MiniGPT4~\cite{zhu2023minigpt-4} to provide an overall caption of the input image. From the generated captions, we extract noun phrases to serve as the semantic tags shared by all the regions in the input image. In addition, we also add an OCR detector~\cite{easyocr} to detect the texts as semantic tags in the scenes.
Note that the generated caption will also be passed to other annotators, which gives visual signal for the LLMs, serving as their eyes. 

\textbf{Imaginator}.
Although the ``spotter'' can find out the major objects in the scenes, it fails to identify many insignificant objects. To address this limitation, we develop an ``imaginator'' to further expand the semantic tag list with plausible imagination.
The ``imaginator'' emulates human-like thinking.
When provided with descriptions of a particular scene, humans can effortlessly imagine the potential objects present. For instance, if informed that an image depicts a group of children standing in a classroom, one may envision objects like ``teacher'', ``blackboard'', and ``stationery''. In our data engine, we utilize Vicuna~\cite{chiang2023vicuna} to imagine possible objects in scenes based on the captions generated by the ``spotter'', and then extend the set using web search engines~\cite{qiu2013visual}. The ``imaginator'' excels at supplementing scene-specific object candidates, such as suggesting ``airport stuff'' instead of simply ``person''. This significantly enhances the concept diversity within this project.

\textbf{Splitter}.
This model is proposed to divide the generated concepts into more fine-grained parts. We find that some region proposals only cover a part of the objects, such as the wing of a plane or the windshield of a car. However, most of the existing perception or caption models are not capable of detecting parts. To this end, we further instruct the Vicuna~\cite{chiang2023vicuna} to divide the semantic tag into parts. For example, ``building" will be decomposed into ``roof'', ``door'', ``windows'' and ``walls''. We tailor the prompt for LLM so that the model only divides the semantic tag that represents a concrete object into parts. LLM is instructed to ignore the semantic candidate that is non-physical or cannot be further divided, such as ``water'', ``sky'', etc. 

\textbf{Magnifier}. 
Although hundreds of open-world semantic tags can be generated by the aforementioned annotators for each image, there still exists some regions whose semantics are absent from the generated tag lists. So we introduce a ``magnifier'' to zoom in on each region and add semantic tags for them.  
We simply crop the region and use a caption model to describe the cropped image, and then extract the noun phrases, which are used as the semantic candidates exclusive for the corresponding regions.
In this model, we use BLIP~\cite{li2022blip} for efficiency.

\subsubsection{Detailed Descriptions} \label{sec:region-desc-gen}

To provide detailed descriptions that include attributes and statuses of each region, we develop a pipeline that expands the region description using the open-world location and its matched semantic tags (see Sec. \ref{sec:matching-semantics} for location-semantic matching). Similar to how we generate semantic tags, we utilize a series of skilled LLMs, including (1) a \textbf{questioner} that asks specific questions about the attributes or status of a given semantic tag; (2) a \textbf{responder} that provides the accurate answers for these questions based on the region's content; and (3) a \textbf{writer} responsible for composing a detailed caption for each region, according to the generated semantic tags, attributes, and status.

\textbf{Questioner}. 
Given semantic tag, to determine its commonly-used attributes,
we use Vicuna~\cite{chiang2023vicuna} as a questioner to generate three questions about the attributes or statuses. The prompt is shown below. In this way, we leverage the world knowledge and reasoning capabilities of LLMs to identify the most relevant attribute of an object.
\begin{center}
\fbox{
\parbox{130mm}{
\textbf{Prompt:} I will give you some objects. Please list 3 questions about the given objects. These questions must be answerable based on a photograph of the object and cannot rely on any outside knowledge. Some examples are listed as follows: \\
\\
\textbf{Human}: Person \\
\textbf{Assistant}: Q1: What is the sex of this person? Q2: What is the hairstyle of this person? Q3: What is this person doing? \\
\\
\textbf{Human}: {\textcolor{blue}{\{Semantic Tag\}}} \\
\textbf{Assistant}:
}
}
\end{center}

\textbf{Responder}. After obtaining the questions related to a semantic tag, we employ Husky \cite{liu2023interngpt}, an LLM-based VQA model, to generate the responses to each question. The responses are generated in several sentences, taking into account the content of the region. An example prompt is shown below. This approach enables us to gather additional information about a region while preventing the inclusion of irrelevant content.
\begin{center}
\fbox{
\textbf{Human}: What is the material of this sphinx?
\textbf{Assistant}:
}
\end{center}

\textbf{Writer}. Based on the question-answering pairs, we proceeded to rephrase them into a single sentence, resulting in a detailed description of the region. The prompt used during annotation is ``Please paraphrase the following sentences into one sentence. \{answer for question 1\} \{answer for question 2\} \{answer for question 3\}''. It is notable that both the question-answering pairs from previous steps and the region captions from this step are valuable for visual recognition and understanding models.

\subsection{Matching Location and Semantic} 
\label{sec:matching-semantics}
Given the generated open-world location and semantic labels, we devise a matching pipeline to select and appropriate tags for each region. Semantic tags that are most related to the region will be picked.

In the matching process, we employ a region-text aligning model to measure the similarity between a certain region and its semantic tag list.
For each region, the semantic tag list is constructed by LLMs (\ie, ``spotter'', ``imaginator'', and ``divider'') and closed-set/grounding object detectors.
Initially, in the first iteration of the data engine, we use a CLIP model~\cite{radford2021clip} for the region-text alignment, where the input is the cropped region. Subsequently, we upgrade the model to our All-Seeing Model. 

In addition, in the first round of data engine, we find that only using CLIP led to erroneous results as it cannot tell which candidate is the major object in the bounding boxes. For example, a bounding box that perfectly frames a person can be classified as a ``backpack'' if the person is carrying a backpack. 
To remedy this, we use CLIPSeg~\cite{luddecke2022clipseg} to generate the mask for each candidate, and the original CLIP confidence is modulated with the corresponding mask area. In this way, the candidate belonging to the main object in the region can be selected. 

\subsection{Human Verification} \label{sec:human-annotation}

Albeit efficient, annotations from the automated pipeline still contains some noise due to the cropping process, which might discard essential context information.  
For instance, a lampshade hanging on the ceiling could be mistakenly described as a ``cup'' due to its similar shape and color.  Therefore, to enhance the data quality, we find it crucial to include human verification.

\textbf{Semantic tags.}
We design a data sampling strategy and simplify the task for annotators by focusing on picking the incorrect ones from the top-5 candidates in each region.
In the real world, concepts exhibit long-tail distribution as shown in Fig.~\ref{fig:semantic-tag-distribution-and-display}. 
Therefore, many rare concepts will be missed if the region is randomly sampled for validation.
To address this issue, we implement a concept-wise sampling strategy. Specifically, we collect a list of concepts in the first 1M images in the \datasetname\ dataset. From this list, we select most concepts for verification. We randomly sample 6 regions from the least frequent concepts and 90 regions from the concepts with the highest number of regions. During the human verification process, the semantic tag list for the sampled regions is provided to the annotators, who are then tasked with filtering out any incorrect tags.

\textbf{Visual Question-Answering Pairs.}
Although using LLMs/VLLMs greatly reduces the annotation cost of generating visual question-answer pairs, there are still some issues that may introduce noise into the data.
(1) The answer to the question is wrong since the VLLM is not perfect.
(2) The generated question for the semantic tag may be unanswerable according to the given image content. 
(3) The semantic tag assigned to a region may be incorrect, leading to meaningless generated questions. For example, if a region containing a dog is wrongly labeled as a cat, asking about the color of the cat would be nonsensical.

To address these issues, we perform a two-stage verification procedure. In the first stage, human annotators are provided with the image, location (bounding box), and corresponding question-answer pairs. They are then asked to annotate the visual question-answer pair with one of four choices: correct answer, wrong answer, unanswerable question, or wrong semantic tag.
Samples annotated as ``correct answer'' are retained, while those annotated as ``wrong answer'' are re-annotated with a correct answer generated by human annotators in the second stage. Samples annotated as ``unanswerable question'' or ``wrong semantic tag'' are annotated with a rejection answer, such as ``This question is unanswerable according to the image'' or ``The object in this region is incorrectly labeled'', respectively.

\textbf{Verification Review.}
We engaged 50 human annotators to perform verification on the annotations generated by our model. To guarantee the quality of this verification process, we additionally request 10 experts to review the verified annotations. These experts are selected based on their domain knowledge and experience in annotation tasks. To streamline the process, we organize the regions requiring review into groups of 100. Each group is assigned to one expert, who checks the accuracy and consistency of the annotations within the group. Any package with an accuracy rate below 95\% will be sent back for re-verification by another annotator. This review process double-checks the annotations, further ensuring their reliability and validity for our models.

\subsection{Data Engine Iteration}

To continuously improve the data quality, we implement a ``data-human-model'' loop that maximizes the utilization of both human-verified data and models. 
As depicted Alg.~\ref{alg:data-engine}, the data engine iteration comprises three steps as follows:
(1) The images are processed with the annotation pipeline which produces automatic annotations.
(2) The ASM model is then trained using these coarse annotations, enabling it to perform both discriminative and generative tasks such as region-text matching and region captioning.
(3) The automatic annotations are sampled and reviewed and corrected by human annotators, yielding high-quality human annotations. This verified data is then used to fine-tune the ASM model, thereby enhancing its performance.
(4) The fine-tuned model is utilized to re-rank the semantic tags and generate more accurate region captions and answers. 
Repeat the third and fourth steps until the data quality meets the requirements.
By following this data iteration process, we ensure continuous optimization of data quality, ultimately leading to superior results.

\begin{algorithm}
	\setstretch{1.2} 
	\renewcommand{\algorithmicrequire}{\textbf{Input:}}
	\renewcommand{\algorithmicensure}{\textbf{Output:}}
	\caption{Data Engine}
	\label{alg:data-engine}
	\begin{algorithmic}[1]
    \REQUIRE \quad  \\ 
        Iteration Number $n$  \\   
    Images $\mathcal{I}$   \\
    Models $\mathcal{M}$  \\
        Annotation Pipeline $P(\mathcal{M}, \mathcal{I})$  \\
    \ENSURE \quad  \\
    Annotations: $\mathcal{A}$  \\
    Improved Models $\mathcal{M}$  \\

        \STATE Generate initial annotation $\mathcal{A}_0$ by off-the-shelf models;
        \STATE Train ASM with $\mathcal{A}_0$, yield $\mathcal{M}_0$;
        \STATE $ i \leftarrow 0 $
        \WHILE{ $ i < n $ }
        \STATE Perform Human verification on $\mathcal{A}_i$, yield $\mathcal{A}_i'$;
        \STATE Fine-tune $\mathcal{M}_i$ with $\mathcal{A}_i'$, obtain $\mathcal{M}_{i+1}$;
        \STATE Obtain Annotation $\mathcal{A}_{i+1}$ by $P(\mathcal{M}_{i+1}, \mathcal{I})$;
        \STATE $i \leftarrow i + 1$
        \ENDWHILE
\end{algorithmic}  
\end{algorithm}

\section{The All-Seeing Model (ASM)} \label{ASM}

\subsection{Overal Architecture}

\emph{Our objective is to create a unified framework that supports contrastive and generative image-text tasks at both the image level and region levels.} 
By leveraging pre-trained LLMs and powerful vision foundation models (VFMs), this model demonstrates promising performance in discriminative tasks like image-text retrieval and zero classification, as well as generative tasks such as visual question answering (VQA), visual reasoning, image captioning, region captioning/VQA, etc.
Additionally, our model shows potential in grounding tasks like phrase grounding and referring expression comprehension, with the assistance of a class-agnostic detector.

\begin{figure}[!t]
	\centering
        \includegraphics[width=1.0\linewidth]{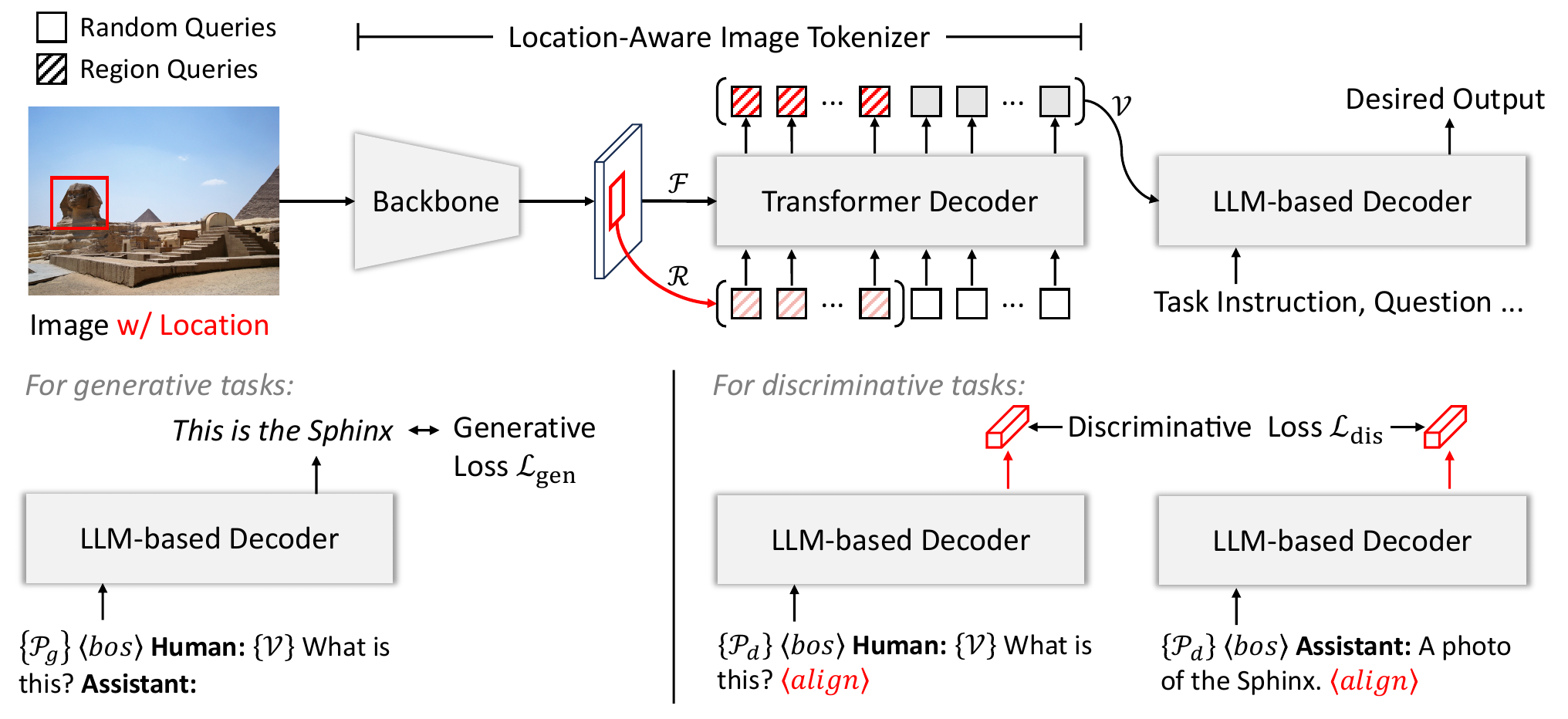}
    \caption{
    \textbf{Architecture and task modeling of the All-Seeing Model (ASM)}. ASM incorporates a location-aware image tokenizer to perform region-text alignment tasks. Image-level and region-level features are encoded as visual tokens $\mathcal{V}$, and fed into the LLM-based decoder along with the users' text input. ASM employs a specific prompt design that allows the LLM decoder to handle both generative tasks and discriminative tasks using a unified architecture with shared parameters. We add soft prompt tokens (\ie, $\mathcal{P}_g$ and $\mathcal{P}_d$) to indicate the desired tasks and use an ``$\left<align\right>$'' token to perform image-text alignment at the LLM's output. $\left<bos \right>$ denotes the beginning token of a sentence.
    }
    \label{fig:asm}
\end{figure}

As illustrated in Fig.~\ref{fig:asm}, our All-Seeing Model (ASM) comprises three key designs:
(1) a \textbf{location-aware image tokenizer} extracting features from both the image and region levels based on the input image and bounding box, respectively.
(2) a \textbf{trainable task prompt} that is incorporated at the beginning of the vision and text tokens to guide the model in distinguishing between discriminative and generative tasks. In the case of the discriminative task, a trainable align token is appended to the input sequence to gather the overall representation, and its embedding is then used in the matching process.
(3) an \textbf{LLM-based decoder} that is utilized to extract vision and text features for discriminative tasks, as well as to auto-regressively generate response tokens in generative tasks.

The training objective of ASM contains two objectives: next token prediction and region-text aligning, as formulated in Eqn.~\ref{eqn:total}. The primary objective focuses on enhancing the model's generation capability, whereas the secondary objective aims to improve its discriminative and retrieval capabilities.
\begin{equation}
\mathcal{L}_{{\rm total}}=\mathcal{L}_{{\rm gen}}+\mathcal{L}_{{\rm dis}}
\label{eqn:total},
\end{equation}
where the generation loss $\mathcal{L}_{{\rm gen}}$ is for the next token prediction, and is the same as the loss of GPT series~\cite{embodiedgpt,radford2019gpt2,brown2020gpt3,openai2023gpt4}. The discriminative loss $\mathcal{L}_{{\rm dis}}$ is for tasks like region-text aligning/retrieval. The discriminative loss follows the contrastive loss of CLIP~\cite{radford2021clip}, where each region is treated as an image when calculating the loss.

\subsection{Location-Aware Image Tokenizer}

To achieve location-aware image tokenizing, we introduce a query-based image tokenizer that conditions its queries on location information, such as bounding boxes, masks, or points.
As depicted in Fig.~\ref{fig:asm}, we first encode the input image using the ViT-g/14 \cite{fang2023eva} backbone, resulting in image features $\mathcal{F} \in \mathbb{R}^{H\times W\times D}$, where $H$ and $W$ denote the spatial size of the feature maps and $D$ denotes the feature dimension.
Next, we use the RoIAlign~\cite{he2017maskrcnn} to extract the region features $\mathcal{R} \in \mathbb{R}^{H_{r}\times W_{r}\times D}$ from the image features $\mathcal{F}$, according to the given bounding box (or mask, point set). Here, $H_{r}$ and $W_{r}$ denote the spatial size of the RoI features.
We then flatten the region features $\mathcal{R}$, use two fully-connection (FC) layers to project them into $\mathcal{Q}_r \in \mathbb{R}^{G \times D_q}$, which has the same shape as randomly initialized query tokens $\mathcal{Q}' \in \mathbb{R}^{G \times D_q}$. Here, $G$ denotes the number of tokens in a query group\footnote{A query group represents a randomly initialized query or a set of query tokens conditioned by a bounding box.}, and $D_q$ denotes the dimension of a query token. 
Subsequently, the $\mathcal{Q}_r$ of $N$ bounding boxes and $\mathcal{Q}'$ are concatenated to form location-aware query tokens $\mathcal{Q} \in \mathbb{R}^{(N + 1)G \times D_{q}}$.
These location-aware query tokens $\mathcal{Q}$ are then passed through a transformer decoder with 12 blocks to extract output features. Finally, the output features are projected to match the feature dimension $D_{t}$ of the LLM and are used as the soft prompt $\mathcal{V}\in \mathbb{R}^{(N+1)G \times D_t}$ for subsequent decoding processes.
Particularly, when no location information is provided, the bounding box is assumed to cover the entire image. This method guarantees a consistent approach for both local region and whole image tokenization.

\subsection{LLM-Based Decoder}  \label{LLM-based Decoder}
To develop a unified LLM-based framework that can effectively handle both generation tasks and discriminative tasks, we utilize Husky-7B~\cite{liu2023interngpt} as our foundation language model to handle various vision-language tasks under the guidance of user instructions and learnable soft prompts that contain image-level and region-level visual information.

\textbf{For generative tasks}, the input sequence comprises three types of tokens, including
(1) learnable generative task prompt $\mathcal{P}_g \in \mathbb{R}^{M\times D_t}$, which informs the model that it should perform a generative task. 
(2) location-aware image tokens $\mathcal{V}$ that contain the extracted image-level and region-level information from the input image and
(3) user prompt that expresses his/her requirements. Given such an input sequence, the LLM generates text tokens sequentially in an autoregressive manner until an end token $\left<eos\right>$ is reached. An example prompt is provided below:
\begin{center}
\fbox{
\textbf{Prompt \#1:}\ \ ``$\{\mathcal{P}_g\}\ \left<bos\right>$\ \textbf{Human: }$\{\mathcal{V}\}$ What is this? \textbf{Assistant:}''
},
\end{center}
where the token number of task prompt $M$ is set to 5. 
$\left<bos\right>$ represents the beginning of the sentence.

\textbf{For discriminative tasks}, different from the CLIP-based framework that directly aligns the output feature from vision and language encoders, we introduce a trainable align token $\left<align\right>$ to extract the holistic representation of the current input sequence. An example prompt for encoding input image is shown as follows:
\begin{center}
\fbox{
\textbf{Prompt \#2:}\ \ ``$\{\mathcal{P}_d\}\ \left<bos\right>$ \textbf{Human:\ }$\{\mathcal{V}\}$ What is this? \textcolor{red}{$\left<align\right>$}''
},
\end{center}
where $\mathcal{P}_d \in \mathbb{R}^{M\times D_t}$ represents the learnable task prompt used for discriminative tasks.

Similarly, the input sequence of input text consists of soft prompt tokens that indicate task information, as well as text tokens that represent the corresponding region caption or object class name. We omit the vision tokens to avoid information leakage. Here is an example prompt:
\begin{center}
\fbox{
\textbf{Prompt \#3:}\ \ ``$\{\mathcal{P}_d\}\ \left<bos\right>$\ \textbf{Assistant:} A photo of the Sphinx.\ \textcolor{red}{$\left<align\right>$}''
}.
\end{center}
During the process of region-text matching, we can achieve image-text retrieval by simply computing the similarity of the embedding of the $\left<align\right>$ token.
It is notable that the learnable task prompt and align tokens used in Prompt \#2 and \#3 are shared, while the task prompt differs between generative tasks (Prompt \#1) and discriminative tasks (Prompt \#2 and \#3).

Compared to the CLIP-based framework, our LLM-based decoder offers two advantages: (1) Our approach builds upon LLMs, allowing us to leverage the powerful world knowledge and reasoning capability of LLMs. (2) Both the image and text aligning embedding of our method are generated by an LLM, which bridges the gap between the pre-training task for the language model and the language-image pre-training task.

\section{Data Analysis}
We conduct an in-depth analysis of our \datasetname\ dataset. We begin by showcasing the abundance of data in terms of quantity. Next, we explore the data diversity and open-world semantics captured in \datasetname. Finally, we thoroughly analyze the data quality of the initial automatic annotation pipeline and explain how we have improved it through data engineering and human feedback.

\subsection{Data Scale}

\textbf{Statistics}.
The \datasetname\ dataset consists of a vast collection of 1.2 billion region-text pairs extracted from 11 million images, encompassing 3.5 million distinct semantic tags. 
Regions in the dataset are categorized into five different resolution scales: tiny, small, medium, large, xlarge, and huge. As indicated in Table~\ref{tab:auto_region_analysis}, the distribution of region resolutions follows a roughly normal distribution. Over half of the regions are on the medium or large scale.
In Sec.~\ref{sec:open-world-localization}, we utilize several region proposal generators, including SAM~\cite{kirillov2023segment}, InternImage~\cite{wang2022internimage}, EVA-02~\cite{fang2023eva02}, and GLIP~\cite{li2022glip}, to generate region proposals for the \datasetname\ dataset.
Table~\ref{tab:region-source-analysis} presents the proportion of regions provided by each model in the 1.2 billion regions. SAM generates 36.4\% of the regions, while the other three models contribute to 63.6\% of the regions. Therefore, although our dataset shares images with SA-1B~\cite{kirillov2023segment} and has a similar number of regions, the actual regions are different due to the use of diverse region proposal generators.

Each region is also annotated with detailed question-answer pairs and a caption, which yields a total of 3.3 billion visual question-answering pairs and 1.2 billion detailed region captions. As seen in Table~\ref{tab:detailed-description-statistics}, the average token number of the answers is $16.91$, while the average token number of the composited caption is $34.84$. The total number of tokens in our detailed region captions amounts to approximately 42.2 billion. This extensive collection of detailed captions provides valuable textual descriptions of regions within the images.

\textbf{Comparisons}.
When comparing the \datasetname\ dataset with popular datasets containing region-level annotations, \datasetname\ stands out with a significantly larger number of regions. 
It has about 33 times more regions than the current largest detection dataset, BigDetection~\cite{cai2022bigdetection}.
While \datasetname\ has fewer images compared to close-set classification datasets~\cite{deng2009imagenet} or vision-language datasets~\cite{schuhmann2022laion5b}, it compensates with valuable region annotations.
Additionally, \datasetname\ offers an abundant collection of detailed region annotations. Compared to the largest region-level dataset, Visual Genome~\cite{krishna2017visualgenome}, \datasetname's detailed region annotation is about 1941 times larger than Visual Genome's 1.7 million pairs of VQA annotations and 222 times larger than its 5.4 million region captions.

\subsection{Data Diversity}
\begin{table}[t]
    \centering
    \small
    \renewcommand{\arraystretch}{1.0}
    \footnotesize
    \begin{tabular}{l|c|c|c|c|c|c|c}
    \toprule
    Region Type & Area Range & Proportion & (V)LLMs & BLIP & InternImage & EVA-02 & GLIP \\
    \midrule
    Tiny & $< 20^2$ & 4.2\% & 33.8\% & 16.5\% & 24.6\% & 25.1\% & 0.0\% \\
    Small & $20^2\sim40^2$ & 8.7\% & 34.5\% & 14.3\% & 24.6\% & 25.9\% & 0.7\% \\
    Medium & $40^2\sim100^2$ & 35.8\% & 55.6\% & 22.9\% & 8.3\% & 11.6\% & 1.7\% \\
    Large & $100^2\sim200^2$ & 23.7\% & 58.5\% & 26.2\% & 5.0\% & 7.9\% & 2.3\% \\
    Xlarge & $200^2\sim500^2$ & 18.3\% & 62.6\% & 27.1\% & 3.0\% & 4.3\% & 3.0\% \\
    Huge & $> 500^2$ & 9.5\% & 69.7\% & 24.9\% & 1.6\% & 1.2\% & 2.7\%\\
    All & $-$ & 100\% & 55.4\% & 24.0\% & 8.2\% & 10.4\% & 2.1\%\\
    \bottomrule
\end{tabular}
    \setlength{\abovecaptionskip}{0.15cm}
    \caption{\textbf{Region statistics and semantic sources}. 
    The percentage of semantic tags generated by different models at each resolution are reported.
    LLM/VLLMs~\cite{chiang2023vicuna,zhu2023minigpt-4,li2022blip} contribute significantly to the semantic diversity of our dataset.}
    \label{tab:auto_region_analysis}
\end{table}

\begin{figure}[!t]
	\centering
        \includegraphics[width=1.0\linewidth]{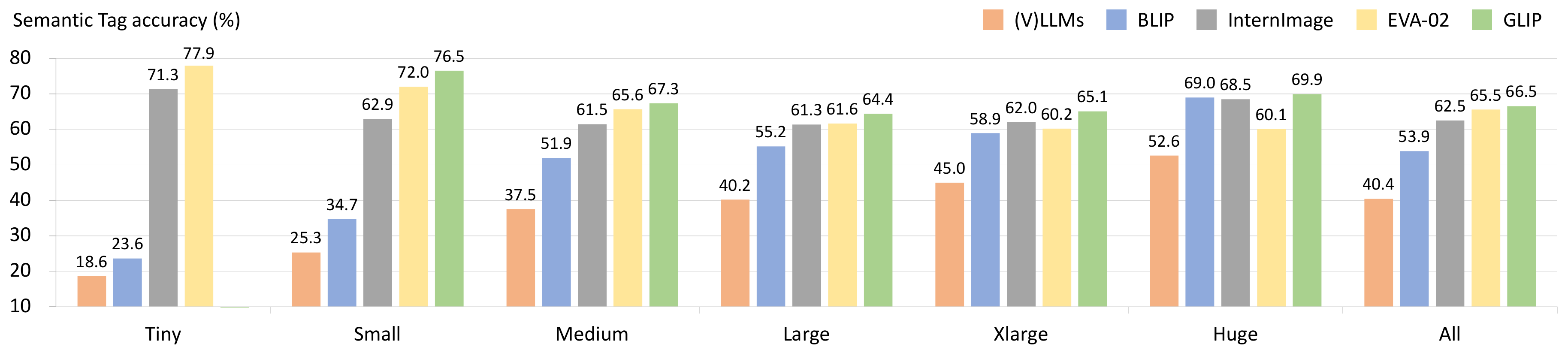}
    \caption{
        \textbf{The accuracy of semantic tags from different sources}.
        LLM/VLLMs~\cite{chiang2023vicuna,zhu2023minigpt-4,li2022blip} show lower accuracy than other models, especially on low resolution regions.
        }
    \label{fig:semantic-tag-accuracy}
\end{figure}

\begin{figure}[!t]
	\centering
	\includegraphics[width=0.99\linewidth]{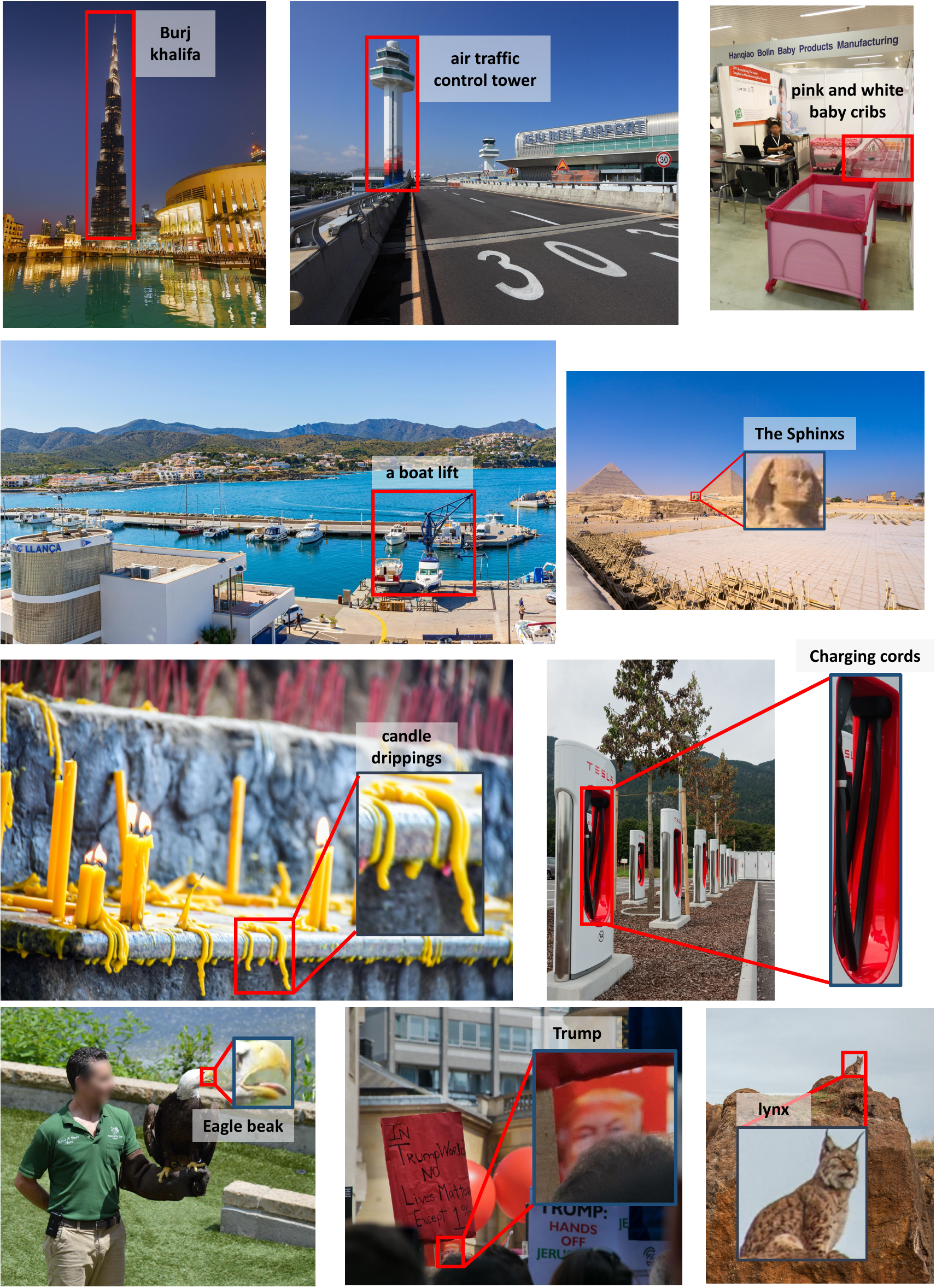}
    \caption{\textbf{Examples of the semantic tags}. Benefitting from the world knowledge of LLMs/VLLMs, the AS-1B dataset covers diversity semantic tags in the real world.}
	\label{fig:brief-region-annotation-examples}
\end{figure}

\textbf{Statistics.} A distinctive feature of \datasetname\ is its vast inclusion of open-world concepts, demonstrated through two key aspects: 1) a large number of semantic tags and 2) long and informative detailed descriptions. Fig.~\ref{fig:brief-region-annotation-examples} visually demonstrates the wide range of open-world concepts present in \datasetname. The dataset covers diverse categories, including fine-grained categories like ``lynx'', proper nouns such as ``The Sphinxs'', object parts like ``charging cords'', and attributes like ``pink and white baby cribs''.
In Fig.~\ref{fig:semantic-tag-distribution-and-display}, we display the frequency distribution of semantic tags, revealing a clear long-tail pattern. The most frequent semantic tags predominantly represent broad category names, while less frequent tags correspond to fine-grained category names or instances with specific attributes. 

In Table~\ref{tab:auto_region_analysis}, we analyze the sources of each semantic tag to understand how open-world concepts are enriched. We report the proportion of sources for the top-1 semantics in the semantic tags at different scales.
The results reveal that 55\% of the top-1 semantic candidates are from the LLM, while 24\% originate from the BLIP (the "magnifier" in Sec.~\ref{sec:open-world-semantics}). Interestingly, only 19\% of the top-1 candidates are generated from the closed-set detectors, InternImage, and EVA-02. This highlights that the majority of concepts in the \datasetname\ dataset are obtained from open-world sources, especially the LLMs and VLLMs.

As for the detailed region caption, the VQA-based generation approach in \datasetname\ has proven advantageous, resulting in longer and more informative region descriptions. A more straight-forward way is to directly ask the VLLM to generate region captions. However, without guidance from semantic tags and questions, the model tends to output inaccurate information or hallucinations.

\textbf{Comparisons}.
Instead of using fixed labels from a pre-defined set, the \datasetname\ dataset employs flexible and open-world semantic tags to label each region. Table~\ref{tab:dataset-comparisons} highlights that \datasetname\ contains a significantly larger number of semantic tags and concepts compared to close-set classification datasets or object detection datasets. For example, the number of semantic tags in \datasetname\ is approximately 159 times greater than the widely-used classification dataset ImageNet-22k~\cite{deng2009imagenet}, and it is 268 times larger than the category number in V3Det~\cite{wang2023v3det}.

\subsection{Data Quality}

\begin{table}[!tb]
	\small
        \setlength{\tabcolsep}{1.9pt}
	\begin{minipage}[t]{0.48\linewidth}
		\centering
		\begin{tabular}{l|cccc}
    \toprule
    Model      & SAM    & InternImage & EVA-02 & GLIP   \\ \midrule
    Proportion & 36.4\% & 20.5\%      & 22.5\% & 20.6\% \\ \bottomrule
    \end{tabular}
    \setlength{\abovecaptionskip}{0.15cm}
    \caption{\textbf{The proportion of region proposals generated by different models}. Only 40\% regions are generated from SAM.}
    \label{tab:region-source-analysis}
	\end{minipage}
	\quad
	\begin{minipage}[t]{0.48\linewidth}
		\centering
		\begin{tabular}{l|ccc}
    \toprule
    Type       & Number & \#Tokens &  Average Tokens \\ \midrule
    Question   & 3.3B           & 34.6B       & 10.50    \\
    Answer     & 3.3B           & 55.4B       & 16.91    \\
    Caption    & 1.2B           & 42.2B       & 34.84    \\ \bottomrule
    \end{tabular}
    \setlength{\abovecaptionskip}{0.15cm}
    \caption{\textbf{The statistics of detailed description in \datasetname\ dataset}. The overall number of tokens reaches 132.2 billion.}
    \label{tab:detailed-description-statistics}
	\end{minipage}
\end{table}

\begin{table}[t]
    \centering
    \small
    \renewcommand\arraystretch{1.0}
    \begin{tabular}{l|cccc}
    \toprule
    Type  & Correct answer  & Wrong answer & Invalid question & Wrong semantic \\ 
    \midrule
    Proportion & 47.1\% & 18.6\% & 19.0\% & 15.3\% \\ 
    \bottomrule
    \end{tabular}
    \setlength{\abovecaptionskip}{0.15cm}
    \caption{\textbf{The statistics of attribute question-answering}. The answers generated by the ``responder" had an accuracy of 47.1\%.
    Wrong semantic denotes that the semantic tags are incorrect.
    }
    \label{tab:detailed-annotation-acc}
\end{table}

\textbf{The Accuracy of Automatic Annotations}.
We evaluated the data quality using two metrics: \textit{top-1 accuracy} and \textit{semantic tag accuracy}. Top-1 accuracy refers to the probability that the top-1 candidates are correct, as selected by the human annotators. On the other hand, semantic tag accuracy denotes the probability the generated semantic tags are selected by the annotators. In the verified annotations, we obtained a top-1 accuracy of $54.8\%$ and a candidate accuracy of $47.0\%$.

As shown in Figure~\ref{fig:semantic-tag-accuracy}, we find that different models in the annotation pipeline exhibit complementary behavior. The LLM and BLIP models show lower accuracy for small regions as they are not robust for the cropped low-resolution images. 
In contrast, close-set detectors perform better on these small regions, providing more accurate semantic candidates. For larger regions, LLMs and VLLMs become more accurate. Hence, the inclusion of close-set detectors can provide a trade-off between data quality and open-world semantics. This interplay of models contributes to the overall improvement of data quality in \datasetname.

As discussed in Sec.~\ref{sec:human-annotation}, the detailed region descriptions are also verified by human experts using a similar procedure. The human annotators are tasked with classifying the VQA pairs into four situations: 1) the question is proper, and the answer is correct; 2) the answer is incorrect; 3) the generated question is unanswerable given the image (\textit{e.g.}, the production date of a car); 4) the semantic tag is wrong. 
As shown in Table~\ref{tab:detailed-annotation-acc}, the accuracy of question-answer pairs is $47.1\%$.

\textbf{Consumption Analysis}. Here we focus on the consumption and efficiency of human verification in the context of the semi-automatic data engine we constructed. This approach significantly reduces the human labor required for data refinement compared with annotating all the data by humans.
For verifying semantic tags, it takes approximately 10 seconds for one annotator to complete one region. Verifying every 1 million regions would take about 2,750 working hours. Considering a group of 50 annotators in our case, the entire verification process takes approximately 15 days. 
If we were to annotate all regions, the annotation consumption would become 1,000 times larger, approximately 42 years. Such a large-scale human annotation effort would be unaffordable.

Moreover, for detailed captions with longer texts, the verification process would take even longer, \textit{e.g.}, 15 seconds for each VQA annotation. Therefore, for large-scale annotation involving billions of regions in our case, utilizing models to annotate data at scale and correcting the models' bias with limited human annotation proves to be both feasible and efficient.

\begin{figure}[!t]
	\centering
        \includegraphics[width=1.0\linewidth]{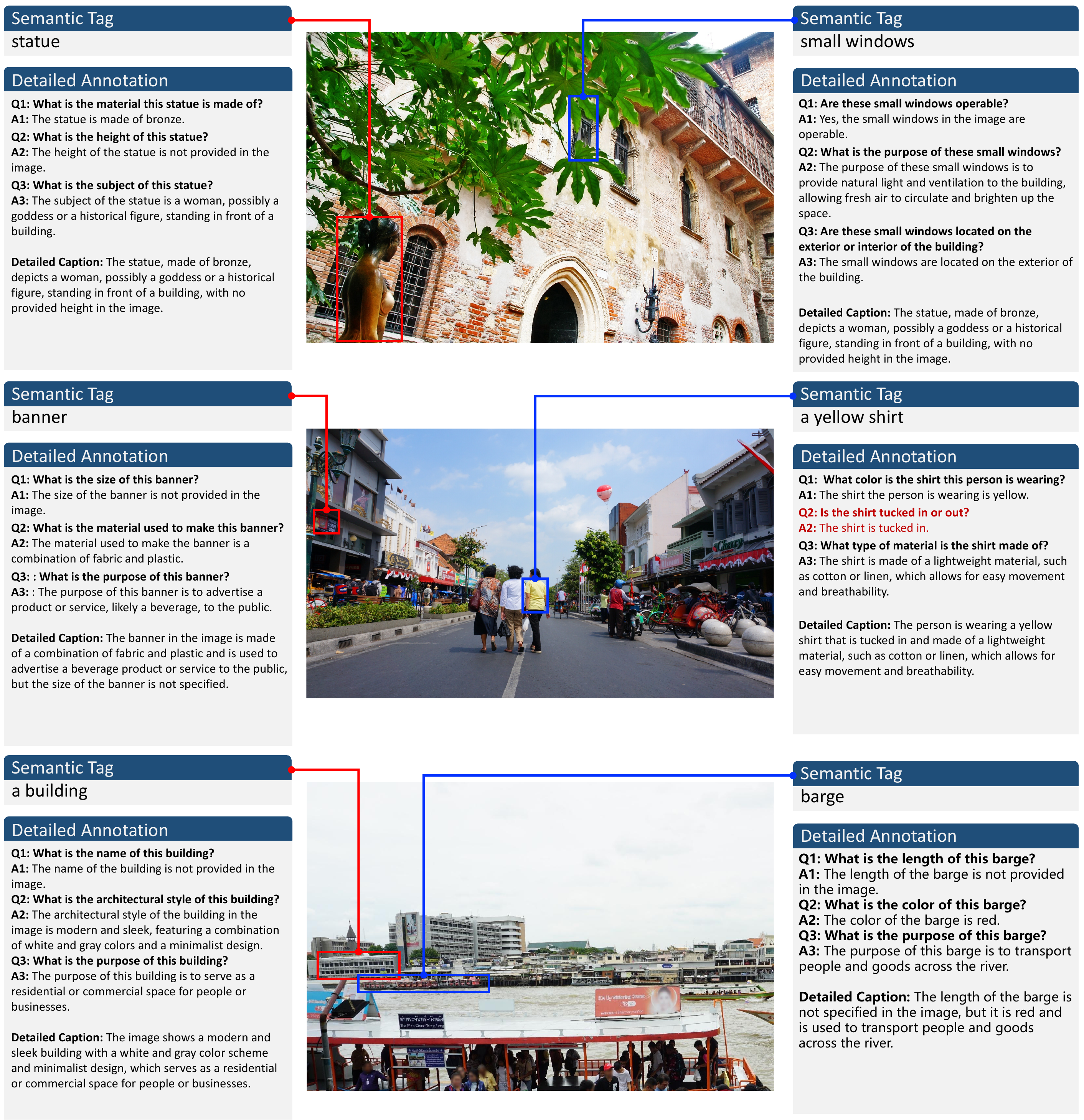}
    \caption{
        \textbf{Examples of the detailed region annotations}. Visual question-answering pairs and captions are provided based on the semantic tags. Failure cases are marked in red.
        }
    \label{fig:detailed-annotation-examples}
\end{figure}

\section{Experiments}

We analyze and compare the proposed ASM with a CLIP-based baseline model and leading
Multi-modality Large Language models (VLLMs) on representative vision tasks including zero-shot region recognition, image-level caption and region-level caption. Additionally, since using conventional image captioning metrics to evaluate LLM-based models can be limiting~\cite{zhao2023chatspot}, we also perform human subject evaluation to compare our model with existing powerful VLLMs~\cite{zhu2023minigpt-4,liu2023llava}.

\subsection{Implementation Details}
\label{sec:implementation-details-and-baselines}
\textbf{Training Setting.}
The training of the All-Seeing Model (ASM) involves three types of labels obtained from the \datasetname\ dataset, including region-level semantic tags, question-answer pairs, and detailed captions. The semantic tags are used for aligning regions with corresponding text, while the other annotations are used to train the text generation task. In addition, we also include LaionCOCO~\cite{laioncoco} in our training process, since the image-level caption data from LaionCOCO is beneficial for ASM's ability to comprehend the whole images.

We adopt a multi-task training approach that combines text generation and region-text alignment tasks to train our ASM.
The batch size for text generation is set to 256, while for region text alignment it is set to 32,768.
We employ the AdamW optimizer~\cite{adamw} with the $\beta_1$ of 0.9, the $\beta_2$ of 0.999, and the weight decay of $0$. During training, the learning rate is initialized as $5\times 10^{-4}$ and includes a linear warmup that lasts until the first 10\% of training steps.
The warmup is followed by a cosine decay strategy with a minimum learning rate of 0.
Unless otherwise specified, the image resolution for all experiments is set to 224 $\times$ 224. We initialize the model parameters using Husky~\cite{liu2023interngpt} and train the model for one epoch.
In addition, we also provide a second-stage fine-tuning setting to further improve the effectiveness of ASM.
Specifically, we utilize high-quality multi-modal data MiniGPT-4~\cite{zhu2023minigpt-4}, LLaVA-150k~\cite{liu2023llava}, and COCO caption dataset~\cite{chen2015coco-caption} as image-level text generation data, along with VG~\cite{krishna2017visualgenome} and RefCOCOg~\cite{mao2016refcocog} datasets as region-level text data. Human-verified region annotations are also included. During fine-tuning, we set the learning rate to $5 \times 10^{-5}$ and apply a weight decay of $0$. The other settings remain the same as during pre-training.
The fine-tuned ASM is denoted as ASM-FT. 

\textbf{Baseline Model}.
To make comparison with recent popular multi-modality large language models (VLLMs)~\cite{zhu2023minigpt-4,liu2023llava, li2023blip-2}
that only focus on processing the entire image, we crop a region from the image and input it to these model for region-level visual recognition and understanding. However, this cropping may result in the loss of some contextual information from the entire image. For better comparison,
we implement a simple region-text contrastive model based on CLIP~\cite{radford2021clip} as a baseline. The baseline model, named Region-Aware CLIP (R-CLIP), is equipped with an RoIAlign layer~\cite{he2017maskrcnn} on the feature maps obtained from the vision encoder in the CLIP model. 
To initialize the model weights, we leverage OpenCLIP~\cite{openclip}~(ViT-L/14) and then train the CLIP model on our \datasetname\ dataset. The model is trained for $10,000$ steps with a batch size of 32,768. Other training settings is the same as those of ASM.

\subsection{Text Generation}
\textbf{Evaluation Setting.}
We evaluate the image-level caption ability of our model on Flickr30K~\cite{flickr30k} and NoCaps~\cite{nocaps} dataset. We report the CIDEr~\cite{vedantam2015cider} and SPICE~\cite{anderson2016spice} metric on these benchmarks. To assess the region-level caption ability, we also evaluate ASM on the Visual Genome~\cite{krishna2017visualgenome} and RefCOCOg~\cite{mao2016refcocog}. On the region caption task, we adopt both the Meteor~\cite{meteor} and CIDEr~\cite{vedantam2015cider} metric as our evaluation metrics. The Meteor, CIDEr, and SPICE metrics are computed by COCOEvalCap\footnote{https://github.com/salaniz/pycocoevalcap}.

\textbf{Results.}
For region-level captioning, as shown in Table~\ref{table:region-caption-automatic-evaluation}, our ASM model surpasses the concurrent region-aware VLLMs, Kosmos-2~\cite{peng2023kosmos2}, by 1.4 points on the RefCOCOg dataset, under the zero-shot setting. After the second-stage fine-tuning, our ASM model has achieved a new record for referring expression generation on RefCOCOg.
Besides, on the Visual Genome (VG) dataset, although the Meteor score of zero-shot ASM is inferior to GRiT~\cite{wu2022grit}, ASM-FT achieves significantly better results than GRiT given relevant data.

In addition, our model also excels at image-level captioning, as presented in Table~\ref{table:global-caption-automatic-evaluation}, our ASM model demonstrates promising zero-shot performance on Flickr30K~\cite{flickr30k} and NoCaps~\cite{nocaps} dataset. Specifically, under the zero-shot setting, our model achieves a CIDEr score of 77.9 without the second-stage fine-tuning and 87.7 after the second-stage fine-tuning, which outperforms all the concurrent VLLMs, such as InstructBLIP~\cite{instructblip}, Shikra-13B~\cite{chen2023shikra} and Kosmos-2~\cite{peng2023kosmos2}. Furthermore, on the NoCaps dataset, ASM also achieves comparable performance compared to the baselines under the zero-shot setting. These results indicate that our ASM model retains a strong image-level comprehension ability while also being region-aware.

In summary, these results highlight the strong region-level text generation capabilities of our model, while also showcasing its ability to comprehend the entire image. The promising zero-shot performance of ASM further demonstrates the effectiveness of our proposed \datasetname\ dataset. Moreover, the unified model structure of ASM enables it to effectively utilize diverse data sources during training, enhancing its overall performance.

\begin{table}[!t]
\centering
\small
\begin{tabular}{@{}l|c|cc|cc@{}}
\toprule
\multirow{2}{*}{Model} & \multicolumn{1}{c|}{\multirow{2}{*}{Zero-shot}} & \multicolumn{2}{c|}{Visual Genome}        & \multicolumn{2}{c}{RefCOCOg}   \\
                       & \multicolumn{1}{c|}{}                           & Meteor        & CIDEr          & Meteor        & CIDEr          \\ \midrule
GRiT~\cite{wu2022grit}                          & \XSolidBrush                                           & 17.1          & 142.0          & 15.2          & 71.6           \\
SLR~\cite{yu2017slr}                            & \XSolidBrush                                           & -             & -              & 15.4          & 59.2           \\
SLR+Rerank~\cite{yu2017slr}                     & \XSolidBrush                                           & -             & -              & 15.9          & 66.2           \\
Kosmos-2 (Few-shot,k=2)~\cite{peng2023kosmos2}   & \XSolidBrush                                           & -             & -              & 13.8          & 62.2           \\
Kosmos-2 (Few-shot,k=4)~\cite{peng2023kosmos2}   & \XSolidBrush                                           & -             & -              & 14.1          & 62.3           \\
Kosmos-2~\cite{peng2023kosmos2}                 & \textcolor{blue}{\Checkmark}                                            & -             & -              & 12.2          & 60.3           \\
\rowcolor{mygray}
ASM                    & \textcolor{blue}{\Checkmark}                                            & 12.6          & 44.2           & 13.6          & 41.9           \\
\rowcolor{mygray}
ASM-FT                 & \XSolidBrush                                           & \textbf{18.0} & \textbf{145.1} & \textbf{20.8} & \textbf{103.0} \\ \bottomrule
\end{tabular}
\setlength{\abovecaptionskip}{0.15cm}
\caption{
\textbf{Performance on the region-level captioning task.} ``-FT'' denotes ASM with second-stage fine-tuning. 
}
\label{table:region-caption-automatic-evaluation}
\end{table}
\begin{table}[!t]
\centering
\small
\begin{tabular}{@{}l|c|rr|rr@{}}
\toprule
\multirow{2}{*}{Model}                              & \multicolumn{1}{c|}{\multirow{2}{*}{Zero-shot}} & \multicolumn{2}{c|}{Flickr30k} & \multicolumn{2}{c}{NoCap}      \\
                                                    & \multicolumn{1}{c|}{}                           & CIDEr          & SPICE         & CIDEr          & SPICE         \\ \midrule
MetaVLM~\cite{hao2022metalm}                        & \textcolor{blue}{\Checkmark}                                            & 43.4           & 11.7          & -              & -             \\
VinVL~\cite{zhang2021vinvl}                         & \textcolor{blue}{\Checkmark}                                            & -              & -             & 95.5           & 13.5          \\
LEMON~\cite{hu2022lemon}                            & \textcolor{blue}{\Checkmark}                                            & -              & -             & 106.8          & 14.1          \\
Flamingo-3B~\cite{alayrac2022flamingo}              & \textcolor{blue}{\Checkmark}                                            & 60.6           & -             & -              & -             \\
Flamingo-9B~\cite{alayrac2022flamingo}              & \textcolor{blue}{\Checkmark}                                            & 61.5           & -             & -              & -             \\
SimVLM~\cite{wang2021simvlm}                        & \textcolor{blue}{\Checkmark}                                            & -              & -             & 110.3          & 14.5          \\
CoCa~\cite{yu2022coca}                              & \textcolor{blue}{\Checkmark}                                            & -              & -             & 120.6          & 15.5          \\
BLIP~\cite{li2022blip}                              & \textcolor{blue}{\Checkmark}                                            & -              & -             & 113.2          & 14.7          \\
BLIP-2~\cite{li2023blip-2}                           & \textcolor{blue}{\Checkmark}                                            & -              & -             & 121.6          & \textbf{15.8} \\
InstructBLIP~\cite{instructblip}                    & \textcolor{blue}{\Checkmark}                                            & 82.8           & -             & \textbf{123.1} & -             \\
Shikra-13B~\cite{chen2023shikra}                    & \textcolor{blue}{\Checkmark}                                            & 73.9           & -             & -              & -             \\
Kosmos-1~\cite{huang2023kosmos-1}                   & \textcolor{blue}{\Checkmark}                                            & 67.1           & 14.5          & -              & -             \\
Kosmos-2~\cite{peng2023kosmos2}                     & \textcolor{blue}{\Checkmark}                                            & 66.7           & -             & -              & -             \\
\rowcolor{mygray}
ASM (ours)                                             & \textcolor{blue}{\Checkmark}                                            & 77.9           & 17.3          & 104.8          & 14.5          \\
\rowcolor{mygray}
ASM-FT (ours)                                             & \textcolor{blue}{\Checkmark}                                            & \textbf{87.7}  & \textbf{18.7} & 117.2          & 15.6          \\ \bottomrule
\end{tabular}
\setlength{\abovecaptionskip}{0.15cm}
\caption{
\textbf{Zero-shot performance on the image-level captioning tasks.} Our ASM shows comparable or even better performance than models dedicated to image-level captioning.
}
\label{table:global-caption-automatic-evaluation}
\end{table}

\subsection{Zero-shot Region Recognition} \label{sec:region-recognition}
\textbf{Evaluation Setting.} We use zero-shot region recognition to evaluate the region-text alignment ability of our model. We use COCO~\cite{lin2014microsoft} and LVIS~\cite{gupta2019lvis} detection dataset for evaluation. Since our current focus is not on object localization, we use the ground-truth boxes and use the model to predict the categories given the corresponding texts following RegionCLIP~\cite{zhong2022regionclip}. We report the mean Average Precision (mAP) metrics for this evaluation.

\textbf{Results.} As shown in Table~\ref{table:clip-automatic-evaluation}, both our baseline model R-CLIP and the proposed ASM achieve promising zero-shot region recognition performance. On the COCO dataset, R-CLIP outperforms the original CLIP by 9.7 mAP, and ASM further increases the performance by 10.4 mAP. On the more challenging LVIS dataset with 1,203 categories, R-CLIP outperforms CLIP by 7.7 mAP, and ASM achieves a more significant improvement of 14.3 mAP over CLIP. These results demonstrate the effectiveness of region-text data in \datasetname\ dataset and the proposed ASM in region-text alignment tasks. Notably, our ASM simultaneously performs caption and region recognition tasks with the same weight, showcasing its versatility and efficiency.

\begin{table}[!t]
\small
\centering
\begin{tabular}{@{}l|cccc|cccc@{}}
\toprule
\multirow{2}{*}{Model} & \multicolumn{4}{c|}{COCO}                                                                                & \multicolumn{4}{c}{LVIS}                                                                                 \\ 
                       & \multicolumn{1}{c}{mAP} & \multicolumn{1}{c}{AP$_{S}$} & \multicolumn{1}{c}{AP$_{M}$} & \multicolumn{1}{c|}{AP$_{L}$} & \multicolumn{1}{c}{mAP} & \multicolumn{1}{c}{AP$_{S}$} & \multicolumn{1}{c}{AP$_{M}$} & \multicolumn{1}{c}{AP$_{L}$} \\ \midrule
CLIP~\cite{radford2021clip}            & 58.9                    & 50.7                     & 70.4                     & 58.3                     & 47.1                    & 40.3                     & 59.2                     & 57.4                     \\
OpenCLIP~\cite{openclip}            & 63.3                    & 47.8                     & 75.6                     & 60.9                     & 49.1                    & 37.4                     & 62.8                     & 66.5                     \\
\rowcolor{mygray}
R-CLIP (our baseline)       & 68.6                    & 61.4                     & 75.4                     & 79.3                     & 54.8                   & 49.3                     & 60.6                     & 66.6                     \\
\rowcolor{mygray}
ASM (ours)                    & \textbf{69.3}                    & \textbf{64.3}                     & \textbf{78.0}                     & \textbf{71.0}                     & \textbf{61.4}                    & \textbf{56.7}                     & \textbf{67.9}                     & \textbf{69.2}                     \\ \bottomrule
\end{tabular}
\setlength{\abovecaptionskip}{0.15cm}
\caption{
\textbf{Zero-Shot object recognition performance.} We report the zero-shot recognition accuracy on COCO and LVIS dataset. The ground-truth boxes are used for inference. 
}
\label{table:clip-automatic-evaluation}
\end{table}

These results demonstrate that, despite the semantic tags in \datasetname\ contain some noise, we can still fine-tune a robust region-aware CLIP model with minor modifications. The result suggests that region-text data in \datasetname\ dataset can be beneficial in enabling the model to learn region-text alignment by considering both the region itself and its context.

\subsection{Data Engineering}
Here, we use quantitative results to show the impact of data quantity and data engineering. Considering the cost of the experiment, we use our baseline model R-CLIP. We use the Zero-shot object recognition metrics as in Sec.~\ref{sec:region-recognition} to inspect the impact of data engineering, \textit{i.e.}, we use the ground-truth boxes and use R-CLIP to determine the categories following RegionCLIP~\cite{zhong2022regionclip}. Unless otherwise specified, we train the model with semantic tags from 1M images in the \datasetname\ dataset.

\textbf{Data Scale up}.
We find that scaling up the semantic tags can be helpful for zero-shot region recognition. To verify this, we train our baseline R-CLIP with different amounts of semantic tags. As shown in Table~\ref{tab:different-amount-of-training-data}, with more training data (from 1M to 5M images), the R-CLIP model attains higher Zero-shot object recognition performance.

\begin{table}[!tb]
	\small
        \setlength{\tabcolsep}{1.9pt}
	\begin{minipage}[t]{0.28\linewidth}
		\centering
		\begin{tabular}{c|cc}
			\toprule
			Data Scale & COCO  & LVIS \\ \midrule
			1M     & 67.8    & 54.0        \\
			2M     & 67.5    & \textbf{55.0}       \\
			5M     & \textbf{68.6}    & 54.8   \\
                \bottomrule
		\end{tabular}
  \setlength{\abovecaptionskip}{0.15cm}
            \caption{\textbf{Zero-shot object recognition performance (mAP)} with different training data scale.} 
		\label{tab:different-amount-of-training-data}
	\end{minipage}
	\quad
	\begin{minipage}[t]{0.28\linewidth}
		\centering
		\begin{tabular}{c|cc}
			\toprule
			Data Cleaning & COCO & LVIS \\ \midrule
			  \XSolidBrush  & 61.8  & 46.5  \\
                \textcolor{blue}{\Checkmark}& \textbf{67.8}  & \textbf{54.0}  \\
   \bottomrule
		\end{tabular}
        \setlength{\abovecaptionskip}{0.15cm}
        \caption{\textbf{Zero-shot object recognition performance (mAP)} with and without data cleaning.}
		\label{tab:different-data-engineering}
	\end{minipage}
 \quad
 \begin{minipage}[t]{0.38\linewidth}
    \centering
    \small
    \begin{tabular}{c|c|cc}
    \toprule
    Human Data & Input Scale &  COCO                             & LVIS \\ \midrule
    \XSolidBrush    &224       & 67.8                                & 54.8       \\
    \textcolor{blue}{\Checkmark} &  224      & \textbf{70.2}  & \textbf{55.0}       \\
    \midrule
    \XSolidBrush        &896   & 76.7                                & 65.7       \\
    \textcolor{blue}{\Checkmark} & 896       & \textbf{80.0}  & \textbf{68.4}     \\ \bottomrule
    \end{tabular}
    \setlength{\abovecaptionskip}{0.15cm}
    \caption{
    \textbf{Zero-shot object recognition performance (mAP)} with and without fine-tuning on human-verified annotations.
    }
    \label{tab:human-data-finetuning}
\end{minipage}
\end{table}

\textbf{Data Cleaning.}
Data cleaning and post-processing are important. In practice, the original data annotation pipeline outputs a total of 2.14 billion regions. We devise a simple data cleaning strategy: (1) we sample the top 100 regions with the highest CLIP score at different scales from each image in the \datasetname\ dataset and (2) we further re-rank the semantic candidates with CLIPSeg~\cite{luddecke2022clipseg}, as discussed in Sec.~\ref{sec:matching-semantics}. This data cleaning process will compress the original 2.14B regions into 1.2B regions. As shown in Table~\ref{tab:different-data-engineering}, adding data cleaning can significantly improve the mAP by $6.0\%$ and $7.5\%$ on COCO and LVIS datasets.

\textbf{How human verification improves the model?}
\label{sec:impact-of-human-verification}
An important part of our data engine is to improve the model with human feedback. In this way, the improved model can be used to refine the initial data which is automatically generated. In this section, we investigate the effectiveness of human verification process. We fine-tune the trained R-CLIP model with human-verified region annotations, and find that a small number of human labels can significantly boost the model performance.

Specifically, to make the most of human labels, we utilized both the positive and negative candidates marked by the human annotators. When calculating the contrastive loss, for each region, we randomly selected one positive candidate and use all the unselected candidates as negative samples. Compared with the image-to-text part in the original CLIP-style contrastive loss, each region will be compared with more negative text samples. The unselected candidates can be viewed as valuable hard samples, indicating when the model will make mistakes. 

\begin{figure}[!t]
	\centering
 \includegraphics[width=0.99\linewidth]{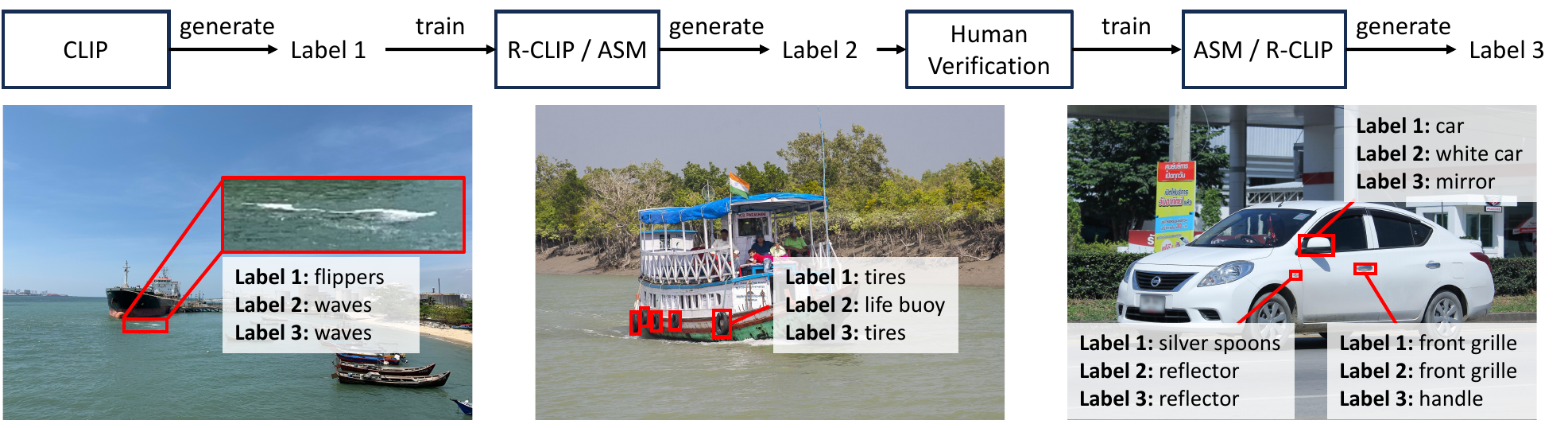}
    \caption{\textbf{Visualization of the data iteration process}. The iteration process improves the label accuracy. We visualize three types of models: (1) \textbf{Label 1}: labels produced the original CLIP; (2) \textbf{Label 2}: labels produced by R-CLIP or ASM, trained with \textbf{Label 1} as input data; (3) \textbf{Label 3}: labels produced by R-CLIP or ASM which is further tuned with human verification data.}
    \label{fig:data-iteration-visualization}
\end{figure}

In practice, we use a batch size of $1024$ and a learning rate of 5e-4 to fine-tune the pre-trained model on the human data for four epochs with only 40k human verified semantic tags. Table~\ref{tab:human-data-finetuning} shows that fine-tuning the model with human data can yield significant performance gain: +2.4 and +3.3 COCO mAP on the resolution of $224$ and $896$. This demonstrates that a small amount of human data can correct the model's bias and hard cases thus improving performance. The effectiveness of human verification lays the foundation for data quality improvement in the data engine iterations.
To intuitively show the data quality improvements, we show the coarse labeling results for CLIP as well as the output of R-CLIP output before and after the human data fine-tuning in Fig.~\ref{fig:data-iteration-visualization}. The original CLIP is unreliable at lower resolutions, \textit{e.g.}, the reflectors and handles on the white cars are categorized into wrong classes. R-CLIP pre-trained on \datasetname\ data performs better in these low-resolution areas. However, it may fail to recognize some objects due to noisy labels, \textit{e.g.}, labeling the tires hung by the boat as a ``life buoy''. The human data fine-tuning process can correct the pre-trained R-CLIP.

\subsection{Human Evaluation}
\label{sec:human-evaluation}
As discussed in ChatCaptioner~\cite{zhu2023chatcaptioner}, using conventional image caption metrics such as Meteor~\cite{meteor} and CIDEr~\cite{vedantam2015cider} may not reliably evaluate relatively lengthy texts generated from LLM-based models. To better assess the text generation ability from a human perspective, we conducted a user study. 

\textbf{Evaluation Setting.} In our user study, we involve a total of 5 participants to evaluate the performance of the All-Seeing Model (ASM) along with two other powerful VLLMs: MiniGPT4~\cite{zhu2023minigpt-4}, and LLaVA~\cite{liu2023llava}. We evaluate image and region-level captioning. For the evaluation, we randomly select 20 samples from each of the Visual Genome, RefCOCOg, COCO, and Flickr30K datasets.
Participants are asked to choose the most informative captions without any factual errors or hallucination. Aside from model outputs, we also add the ground truth captions as options, which can be viewed as human outputs.

\textbf{Results.}
The human evaluation results in Table~\ref{table:captaion-human-evaluation} indicate that captions generated by our ASM are preferred over those from MiniGPT4 and LLaVA. While LLaVA and MiniGPT4 may produce longer captions for region-level tasks (VG and RefCOCOg), they often introduce over-association, hallucinations, and factual errors. In contrast, ASM generates captions with moderate length and more accurate information. On RefCOCOg, Flickr30K, and NoCaps datasets, ASM even outperforms human annotations with longer and more detailed captions. This is because human annotators tend to write short captions, while users prefer longer, detailed captions generated by ASM, which also contain fewer factual errors. For image-level generation tasks, ASM produces captions with similar length to those from MiniGPT4 and LLaVA but is more frequently favored by users.

The results clearly demonstrate the effectiveness of ASM and the AS-2B dataset. The VQA-based annotation pipeline provides region-specific information with less irrelevant content, reducing the occurrence of hallucinations. Moreover, human verification further enhances the data quality, leading to significantly better performance on region-level tasks.

\begin{table}[!t]
\centering
\small
\begin{tabular}{l|cc|cc|cc|cc}
\toprule
Model      & \multicolumn{2}{c|}{Visual Genome} & \multicolumn{2}{c|}{RefCOCOg} & \multicolumn{2}{c|}{Flickr30K} & \multicolumn{2}{c}{NoCaps} \\
           & Rate    & Length   & Rate       & Length      & Rate         & Length        & Rate       & Length       \\ \midrule
Human                               & 47.8    & 13.6             & 10.3       & 6.3                   & 30.0       & 16.0                 & 27.3         & 15.1                  \\
LLaVA~\cite{liu2023llava}           & 4.3     & 110.8            & 15.4       & 100.6                 & 17.5       & 114.0                & 9.1          & 108.4                 \\
MiniGPT4~\cite{zhu2023minigpt-4}    & 8.7     & 110.9            & 15.4       & 113.5                 & 14.2       & 114.6                 & 13.6         & 101.0                  \\
\rowcolor{mygray}
ASM (ours)                          & 39.2    & 37.5             & 46.1       & 33.6                  & 38.3       & 112.4                & 50.0         & 102.1                 \\ \bottomrule
\end{tabular}
\setlength{\abovecaptionskip}{0.15cm}
\caption{
\textbf{Human evaluation results on caption tasks.} We ask the users to select the caption that contains the most information regarding the image/region while does not producing any factual errors.
}
\label{table:captaion-human-evaluation}
\end{table}

\section{Conclusion}
In this paper, we present the All-Seeing (AS) Project, which develops a comprehensive system for panoptic visual recognition and understanding in the open world from both dataset and model perspectives. 
In terms of data, we elaborate a semi-automatic data engine consisting of an automatic annotation pipeline and a human verification step. Using this data engine, we annotated the \datasetname\ dataset comprising over 1 billion region-level comprehensive annotations, with controllable costs. 
From the model aspect, we propose a region-aware multi-modal large language model called the All-Seeing Model (ASM). The ASM utilizes a unified LLM decoder to model both region-text alignment and image-conditioned text generative tasks. 
Leveraging the \datasetname\ dataset and other high-quality data, ASM achieves state-of-the-art results on image and region-level tasks. We also fine-tune a region-aware CLIP model exclusively on the \datasetname\ dataset, surpassing the original CLIP by significant margins in region recognition. We believe that the data engine, \datasetname\ dataset, and the ASM model proposed in the All-Seeing Project will inspire further research and development towards empowering artificial intelligence systems with an ``all-seeing eye,'' enabling them to achieve a deeper understanding of the world.

\section*{Credit Attribution of Equal Contribution Authors}

\textbf{Weiyun Wang} is responsible for the implementation and experiments of ASM, constructing the detailed annotation data pipeline, optimizing the reasoning efficiency of LLM/VLLM-related annotation modules, refactoring the code of R-CLIP and improving its efficiency, implementing the code of the open-world semantic generation, and drafting the corresponding method and experiment sections.

\textbf{Min Shi} is responsible for managing the construction of the data engine, joint debugging the data engine,
constructing the semantic tag annotation pipeline, designing data cleaning and conducting data-related analysis and experiments, implementing part of the R-CLIP's code and main experiments of R-CLIP, implementing the open-world semantic matching, participating in the human verification process, drafting partial manuscripts and revising the manuscript.

\textbf{Qingyun Li} is responsible for the main part of the open-world localization, optimizing partial localization models, implementing the main code of the R-CLIP, some refining experiments, setting up the human evaluation platform for ASM, and drafting partial manuscripts.

\textbf{Wenhai Wang} is the technical manager of the AS project, responsible for the task decomposition, prototyping, and optimization suggestions of each part of the project, and drafted and revised the entire manuscript.

\textbf{Zhenghang Huang} is responsible for the main part of the human verification process, setting up the human verification platform, implementing part of the location annotators, communicating and guiding the manual annotation team, and drafting partial manuscripts.

\textbf{Linjie Xing} is responsible for optimizing most of the location annotator and part of the semantic generation modules, implementing part of the location annotators, reviewing part of the human verification results, and drafting partial manuscripts.

Special acknowledgment to Xizhou Zhu and Hao Li for the preliminary idea and verification of the AS project.

\clearpage

{
\small
\bibliographystyle{plain}
\bibliography{egbib}
}

\end{document}